\documentclass{article}
\usepackage[preprint]{unireps_2026}

\usepackage[utf8]{inputenc} % allow utf-8 input
\usepackage[T1]{fontenc}    % use 8-bit T1 fonts
\usepackage{hyperref}       % hyperlinks
\usepackage{url}            % simple URL typesetting
\usepackage{booktabs}       % professional-quality tables
\usepackage{amsfonts}       % blackboard math symbols
\usepackage{nicefrac}       % compact symbols for 1/2, etc.
\usepackage{microtype}      % microtypography
\usepackage{xcolor}         % colors

\usepackage{graphicx}
\usepackage{amsmath}
\usepackage{amssymb}
\usepackage{upquote}
\usepackage{array}
\usepackage{natbib}
\usepackage{wrapfig}
\newcommand{\dmu}{\Delta\mu}
\newcommand{\SigW}{\Sigma_{W}}
\newcommand{\KL}{\mathrm{KL}}
\newcommand{\wlda}{w_{\mathrm{LDA}}}

\title{Do Probing and Steering Favor the Same Directions Within and Across Language Models?} %Evidence Within and Across Model Families

\author{
  Afshin Sadeghi\thanks{Related code and interactive paper page in http://afshn.com/research/probing-vs-steering} \\
   AI Assurance and Assessments (AAA) \\
  Fraunhofer IAIS, Sankt-Augustin, Germany\\
 \texttt{afshin.sadeghi@iais.fraunhofer.de} \\
  \And
  Daniel Becker \\
AI Assurance and Assessments (AAA) \\
  Fraunhofer IAIS, Sankt-Augustin, Germany\\
  \texttt{daniel.becker@iais.fraunhofer.de} \\
}

\begin{document}

\maketitle

\begin{abstract}
A direction in a language model's hidden state can be used to \emph{probe} whether
a concept is present or to \emph{steer} the model toward it. Although probe
directions are often reused for steering, the direction that best detects a concept
need not be the one that best changes it. We study this distinction in three
instruction-tuned models from different families. We construct a one-parameter
family of directions ranging from a covariance-adjusted discriminant to the mean
difference, and evaluate probing and additive steering, which adds the direction to the hidden state, along the family at a fixed side
effect. For truthfulness and sentiment, in every model, the best probe is not the best steering direction: steering consistently favors directions closer to the mean difference.
This separation holds across 59 of 60 layer--model--concept combinations in a layer sweep, with matched probe and steering texts in all six model--concept combinations, and under a rotation-based steering operator. The exact steering optimum is less stable: for sentiment, an intermediate direction produces a 27--56\% stronger steering effect than the mean difference, but this gain shrinks or disappears with matched texts and depends on the layer and the operator.

The two roles also transfer differently across models. A probe transferred by a linear map fitted on neutral text retains 81--91\% of the target model's above-chance probing performance, while the transferred mean difference retains less of its steering effect in 11 of 12 model pairs. This is a consistent tendency, although
our predefined criteria classify only three pairs, and nine remain inconclusive. Finally, as an application on PopQA, steering toward ``unknown'' increases abstention by 18--44 percentage points on uncertain errors, versus 2--5 on confident errors, while rarely correcting errors in cases where a correct answer is reachable.
\end{abstract}

\section{Introduction}
\label{sec:intro}

A language model's hidden state can be used, without changing the model's weights, to \emph{probe} for a concept or to \emph{steer} the model toward it.
A direction can predict a concept well because it exploits variation correlated with the concept, without being the direction along which the model is most effectively changed.
Steering methods nevertheless often reuse a probe direction, for example, the reading vector of representation engineering \citep{zou2023repe}.
\citet{park2024linear} relate the representations used for probing and for steering through a causal inner product, which raises an empirical question: does the direction that probes a concept best also steer it best in the ordinary geometry of the hidden state?
Earlier comparisons put a trained probe against the mean difference, two fixed directions, and find that the mean difference can steer better \citep{li2023iti,marks2024geometry} and that the two can detect equally well but intervene differently \citep{wu2025axbench}.
\begin{wrapfigure}{r}{0.4\linewidth}
%\vspace{-2pt}
\centering
\includegraphics[width=\linewidth]{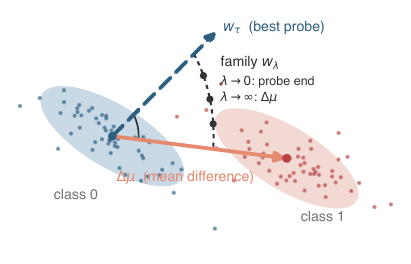}
\caption{Hidden states of one layer of LLM $A$. $\Delta\mu$ joins the two class means; the best probe $w_\tau$ is tilted away from it by the within-class covariance, and the family $w_\lambda$ runs between them. Probing reads $w^{\top}h$; steering adds $a\|h\|w$. The paper also asks how these two transfer across language models.}
\label{fig:concept}
\vspace{-12pt}
\end{wrapfigure}

In contrast to these two-endpoint comparisons, which show which of the two directions works better for each operation at a chosen steering strength, we ask where between them each operation works best, at a fixed side effect on unrelated text, so that no direction gains only by perturbing the model more.
We therefore construct a continuous family of directions that extends from a covariance-adjusted discriminant to the mean difference, the direction commonly used for steering, and we measure probing and steering along the whole family at a fixed side effect, in three model families, and carry each role across models under its own transfer law (Figure~\ref{fig:concept}).

The distinction matters most when representations are compared across models.
Recent work reports convergence between independently trained models \citep{huh2024platonic}, but whether two models agree on a concept may depend on whether the concept is read or changed.
A linear map between two models carries a steering direction forward, whereas a probe has to be carried by the transpose of the reverse map (Section~\ref{sec:transfer}).
Each role can therefore be transferred under its own law, which turns representational alignment into a functional question: do models agree more on how a concept is read than on how it is changed?

%As an application, we use the distinction on factual questions, to separate wrong answers whose correct answer is reachable from those without one, and to steer toward abstention.

Our two main results concern the split between the roles.
First, along a continuous family of directions and at a fixed side effect, the direction that probes a concept best is not the one that steers it best. This holds in all six combinations of model and concept at the main layer, also with matched probe and steering texts, and, in separate analyses, at 59 of 60 combinations of layer, model and concept for truthfulness and sentiment in a layer sweep and in the four cases where a rotation-based operator reaches the same side effect (Appendices~\ref{app:layers}, \ref{app:matched} and~\ref{app:rotation}).
Second, between models of three families, each role carried under its own law, probes keep most of their performance while the mean difference keeps less of its steering effect in most pairs.
A secondary result is that where steering works best, and whether an intermediate direction beats the mean difference, depends on the concept, the layer, the operator, and the texts.

As an application, using an error taxonomy adapted from \citet{orgad2025llms}, we show that steering toward ``unknown'' turns uncertain errors into abstentions far more readily than confident ones, while the probe and steering directions for this concept are less clearly separated than for truthfulness and sentiment, and gating by the probe protects correct answers but passes our test only in Qwen.

\section{Method: Probing, Steering, and Transfer}
\label{sec:method}
Two research questions outline this study.
We ask whether probing and steering favor the same directions, within a model and across models (RQ1), and whether this distinction helps to separate wrong answers by whether a correct answer is reachable and to steer the model toward abstention (RQ2).
Section~\ref{sec:directions} defines the family of directions and how probing, steering, and their side effect are measured on it; Section~\ref{sec:transfer} describes how directions are carried between models; Sections~\ref{sec:rq1} and~\ref{sec:rq2} present the hypotheses that the experiments evaluate.

\subsection{Background and Definitions}
\label{sec:directions}

\paragraph{Probe score.}
We treat each concept as binary and use texts labeled for its presence ($y=1$) or absence ($y=0$), for example, true TruthfulQA answers and positive Yelp reviews labeled $y=1$.
Let $h\in\mathbb{R}^d$, with $d$ the hidden size, be the hidden state at the last token of the text after a fixed layer $\ell$ (Section~\ref{sec:models-data}).
This is the residual-stream vector at that token, the sum of the token's input embedding and the outputs of all attention and feed-forward blocks up to and including layer $\ell$.

We define a probe as a direction $w$ in the space of $h$, and its probe score as $w^\top h$.
Appendix~\ref{app:pool} shows that reading $h$ at the last token separates the labels better than max pooling over the tokens, the reading used by \citet[Sec.~3.2]{wu2025axbench}, for both concepts and all three models.
Below we define the family of directions and compute each direction once, in closed form, from the $(h,y)$ pairs of an estimation subset.

\paragraph{Mean difference.}
Let $\mu_1=\mathbb{E}[h\mid y=1]$ and $\mu_0=\mathbb{E}[h\mid y=0]$ be the mean representations of the two classes. The mean difference $\dmu=\mu_1-\mu_0$ points from $\mu_0$ to $\mu_1$ and ignores how much the representations vary within each class; its normalized form, $\dmu/\|\dmu\|$, is the $\lambda\to\infty$ end of the family defined below.

Up to a scale factor, $\dmu$ is the steering vector of contrastive activation addition (CAA) \citep{rimsky2024caa}, so our comparisons with $\dmu$ are also comparisons with CAA; it is also the refusal direction \citep{arditi2024refusal}, a strong AxBench baseline \citep{wu2025axbench}, and the target of Spherical Steering \citep{you2026spherical}. Related methods use a single prompt pair \citep{turner2023activation} or a principal component of differences \citep{zou2023repe}.

\paragraph{Within-class covariance.}
We measure how much the representations vary within each class by the pooled within-class covariance $\SigW=(1-\pi)\,\Sigma_0+\pi\,\Sigma_1$, where $\Sigma_0$ and $\Sigma_1$ are the covariances of $h$ within the classes $y=0$ and $y=1$, and $\pi=\Pr(y=1)$ is the class prior.
Directions in which $\SigW$ is large are directions along which the examples of one class spread widely.

\paragraph{Fisher discriminant.}
A probe's score should separate the two classes, which a direction of large mean separation does poorly if the representations also vary strongly along it within each class.
For the other end of the family we therefore want the direction that separates the classes best and, like $\dmu$, has a closed form.
For nonsingular $\SigW$, we choose the Fisher discriminant $\wlda\propto\SigW^{-1}\dmu$, which adjusts the mean difference by the within-class covariance and maximizes the ratio of between-class to within-class variance of the probe score $w^\top h$.

Under a shared-covariance Gaussian model, $\wlda$ is the direction of the Bayes-optimal classifier, so no other score separates the classes with a higher area under the ROC curve (Appendix~\ref{app:witness}).
Logistic-regression probes share this population direction for such classes; with $n<d$ and separable classes, however, a trained logistic probe tends toward the maximum-margin direction as its penalty vanishes, so our results concern the closed-form family rather than trained probes.

\paragraph{A regularized family.}
The matrix $\SigW$ has one row and one column for each of the several thousand dimensions of $h$, but it is estimated from fewer examples and is therefore singular, with no inverse.
We therefore use the regularized family
\begin{equation}
w_{\lambda}
=
\frac{(\SigW+\lambda I)^{-1}\dmu}
{\|(\SigW+\lambda I)^{-1}\dmu\|},
\label{eq:wlambda}
\end{equation}
where $\lambda>0$ controls how strongly the covariance structure is taken into account and $I$ is the identity matrix.
As $\lambda\rightarrow\infty$, $(\SigW+\lambda I)^{-1}\dmu\approx\lambda^{-1}\dmu$, so $w_\lambda\rightarrow\dmu/\|\dmu\|$, and as $\lambda$ decreases, the covariance adjustment grows.
For nonsingular $\SigW$, the limit $\lambda\rightarrow0$ recovers $\wlda$; for singular $\SigW$, it is the component of $\dmu$ outside the span of the within-class deviations, not the Fisher discriminant (Appendix~\ref{app:witness}). We report $\lambda$ in units of $\tau=\operatorname{tr}(\SigW)/d$, where $\operatorname{tr}$ is the trace; $\tau$ is thus the average within-class variance per dimension of $h$.
Raw values of $\lambda$ cannot be compared across models, because $\tau$ is about $0.01$ in LLaMA and $4\,000$--$7\,000$ in Gemma. At $\lambda=\tau$, $\SigW$ and a spherical covariance with the same total variance receive equal weight in every model (Appendix~\ref{app:witness}).

We call $w_\lambda$ a regularized discriminant direction, because its probe score $w_\lambda^\top h$ is meant to separate the two classes, and we call the small-$\lambda$ and large-$\lambda$ regions of the family its probe end and its mean-difference end.

\paragraph{Probing, steering, and side effect.}
The same unit vector $w_\lambda$ is used in two ways, and only $\lambda$ changes it.
Used as a probe, a direction is judged by its probing performance $R$, the area under the ROC curve (AUROC) of its probe score on held-out examples, where $0.5$ is chance.
Used for steering, $w$ enters the additive intervention $T(h;w,a)=h+a\|h\|w$, applied with strength $a$ from the last token of a steering prompt onward, for truthfulness, including the answer tokens whose likelihood is scored (Appendix~\ref{app:data}); the resulting change of a behavioral measure is the steering effect $S$.

A stronger push changes more than the target concept, and we therefore compare directions at a fixed side effect
\begin{equation}
K=\frac{1}{|N|}\sum_{x\in N}
\KL\!\left(p_0(\cdot\mid x)\,\|\,p_T(\cdot\mid x)\right),
\label{eq:K}
\end{equation}
where $N$ is a set of neutral sentences and $p_0(\cdot\mid x)$ and $p_T(\cdot\mid x)$ are the model's next-token distributions after the sentence $x$ without and with the intervention $T$.
$K$ is the average Kullback--Leibler divergence in nats, and $K=0$ means that the intervention does not change what the model predicts on neutral text.
Every direction's steering effect is read at the same budget $K=\kappa_0$, a small value fixed separately for each model and concept (Section~\ref{sec:eval}).
This controls the size of the intervention only on neutral text; equal $K$ need not imply equal disruption of task-specific behavior.

\paragraph{Why probing and steering can prefer different directions.}
Measuring $R(\lambda)$ and $S(\lambda)$ along the family asks whether probing and steering prefer the same direction; if they favor different regions, predictive information and causal steerability are not optimized by the same direction.
The two ends of the family show why this can happen.
For a linear probe $w$, the corresponding activation pattern is proportional to $\Sigma_h w$, where $\Sigma_h$ is the covariance of $h$ \citep{haufe2014interpretation}.
In the population, taking $\wlda=\SigW^{-1}\dmu$ (its scale does not matter below),
\begin{equation}
\Sigma_h\wlda
=
\left(1+\pi(1-\pi)\dmu^\top\SigW^{-1}\dmu\right)\dmu
\propto\dmu .
\label{eq:pattern}
\end{equation}

Thus, the probe's filter $\wlda$ and the direction $\dmu$ along which the classes differ are different objects; Appendix~\ref{app:witness} gives the statistical interpretation of the two ends.

\subsection{Probing and Steering Across Models}
\label{sec:transfer}

Let $h_A(x)\in\mathbb{R}^{d_A}$ and $h_B(x)\in\mathbb{R}^{d_B}$ be the hidden states of models $A$ and $B$ on the same input $x$, defined as in Section~\ref{sec:directions}.
A direction of $A$ cannot be used in $B$ directly, because the two states lie in spaces with different dimensions and bases.
We therefore connect the two spaces with an affine map $h_B(x)\approx M_{A\rightarrow B}\,h_A(x)+b_{A\rightarrow B}$, fitted on neutral text without concept labels, where $M_{A\rightarrow B}\in\mathbb{R}^{d_B\times d_A}$ and $b_{A\rightarrow B}\in\mathbb{R}^{d_B}$ are estimated by ridge regression of $h_B$ on $h_A$ (Appendix~\ref{app:derive}).
The reverse map, $M_{B\rightarrow A}$ and $b_{B\rightarrow A}$, is fitted in the same way with the two models exchanged; it is not the inverse of $M_{A\rightarrow B}$, which does not exist when $d_A\neq d_B$.

A steering direction is a displacement of the hidden state, and the offset cancels when a displacement is mapped, so we transfer it forward, $\delta_B=M_{A\rightarrow B}\,\delta$, and use $\delta_B/\|\delta_B\|$ as the unit direction $w$ in the intervention $T$ of model $B$, so that the strength $a$ keeps its meaning.
A probe is a linear functional, which turns a hidden state into a score, and writing the state of $A$ through the reverse map shows that a probe is transferred by the transpose of the reverse map, $w_{A\rightarrow B}=M_{B\rightarrow A}^{\top}w_A$ (Appendix~\ref{app:derive}).
We call these the steering law and the probe law.
They are those of \citet[Sec.~4, App.~B.2]{chen2025stitching}, who train the maps jointly so that they invert each other; we fit each map separately in closed form, and what we add is the comparison of how well each role transfers under its own law.

Because the maps never see the concept texts, a transferred direction can work in the target only to the extent that both models encode the concept along directions that the map puts into correspondence; the transfer thus tests what a direction does in the target, not whether the two representations are identical.
To check that the choice of law matters, we also use an orthogonal Procrustes map as a control, for which the probe law and the steering law give the same vector (Appendix~\ref{app:figs-xmodel}, Table~\ref{tab:xmodel-controls}).

\subsection{RQ1: Probing versus Steering}
\label{sec:rq1}
We test RQ1 with three hypotheses: H1a and H1b compare probing and steering within each model, and H1c compares them across models.
Each hypothesis names what we measure and the rule, fixed before the experiments, by which it holds or fails (Appendix~\ref{app:rules}).

\paragraph{Within models.}
Within each model, we measure $R(\lambda)$ and $S(\lambda)$ on the same held-out questions and reviews, which the probe reads as labeled texts and steering uses as prompts (Section~\ref{sec:eval}).
We call the plateau of a curve the set of directions whose value lies within one standard error of its best value.
H1a asks whether the steering plateau lies at or beyond the probing plateau toward $\dmu$; the claim is universal only if it holds in all three models.

If some $\lambda$ steers better than $\dmu$, H1b tests whether this gain carries over to new data.
Let $S_{(1)}$ and $S_{(2)}$ be the steering effect on the first and the second held-out split.
We choose $\lambda^*=\arg\max_\lambda S_{(1)}(\lambda)$ on the first split, and H1b holds if $S_{(2)}(\lambda^*)\ge1.1\,S_{(2)}(\dmu)$ and the paired 95\% bootstrap interval for the difference excludes zero.
This guards against selecting on the test data, but for truthfulness the directions are computed and evaluated on TruthfulQA, so a gain from the labels cannot be separated from one from matching the test distribution.

\paragraph{Across models.}
H1c compares probes and steering directions transferred from $A$ to $B$:
\begin{equation}
\rho_{\mathrm{probe}}
=
\frac{
R_B(M_{B\rightarrow A}^{\top}w_{A,\hat\lambda})-\tfrac12
}{
R_B(w_{B,\hat\lambda})-\tfrac12
},
\qquad
\rho_{\mathrm{steer}}
=
\frac{
S_B(M_{A\rightarrow B}\dmu_A)
}{
S_B(\dmu_B)
},
\label{eq:rho}
\end{equation}
where $R_B$ and $S_B$ are the probing performance and the steering effect in $B$, $\hat\lambda$ is the target's best probing $\lambda$ on the grid of multiples of $\tau$, the source direction uses the same multiple of its own $\tau$, $\lambda_A=(\hat\lambda/\tau_B)\,\tau_A$, and $S_B$ is read at $B$'s own budget $\kappa_0$.

$\rho_{\mathrm{probe}}$ compares the probe of $A$, carried into $B$ by the probe law, with $B$'s own probe at the same $\lambda$.
Because an AUROC of $\tfrac12$ is chance, we subtract $\tfrac12$ from both AUROCs, so $\rho_{\mathrm{probe}}$ is the share of the target probe's above-chance AUROC that the transferred probe keeps, and a transferred probe at chance gives $0$.
$\rho_{\mathrm{steer}}$ compares the mean difference of $A$, carried into $B$ by the steering law, with $B$'s own mean difference.
For both ratios, $1$ means as good as the target's own direction, $0.5$ half as good, and values above $1$ are better.
We fix the thresholds before evaluation: $\rho\ge0.8$ denotes strong sharing and $\rho<0.5$ weak sharing; they are round values rather than estimates, so we also show the ratios themselves.
Pairs are classified as probe-shared, steering-shared, both-shared, neither-shared, or inconclusive.

\subsection{RQ2: Answer Reachability and Abstention}
\label{sec:rq2}

We study wrong answers through the same split between probing and steering, and separate them by the model's own answers to the same question.
For each PopQA question (Section~\ref{sec:models-data}), with ``I don't know'' allowed, the model gives one greedy answer and five sampled answers, each judged correct, wrong, or an abstention.
C0 is the set of questions whose greedy answer is correct.
C1 is the set of questions whose greedy answer is wrong although at least one sample is correct, that is, errors for which a correct answer is reachable under our decoding; we call them \emph{knowledge-available errors} for short.
C2, the \emph{uncertain errors}, and C3, the \emph{confident errors}, are the sets of questions with no correct answer among the six whose wrong greedy answer is repeated in at most two or in at least four of the five samples; Cm holds the remaining such errors, repeated in exactly three samples, and AK and AU the greedy abstentions with and without a correct answer among the six (Table~\ref{tab:cases} in Appendix~\ref{app:rq2}).

The probe of RQ2 is a member of the family of Section~\ref{sec:directions}, fitted to separate questions with and without a correct answer among the six and read at the last prompt token, before any answer is generated.
These labels describe whether a correct answer is reachable under the decoding we test, not whether the model knows the fact or recognizes its own error.

We test RQ2 with three hypotheses.
H2a (probing) compares the probe with an output-only baseline, i.e., the entropy of the first answer token, $H_{\mathrm{ent}}(q)=-\sum_v p(v\mid q)\log p(v\mid q)$.
It asks whether the hidden state separates (i) knowledge-available errors from errors without a correct answer (C1 vs.\ $\text{C2}\cup\text{Cm}\cup\text{C3}$, and, with abstentions included, $\text{C1}\cup\text{AK}$ vs.\ $\text{C2}\cup\text{Cm}\cup\text{C3}\cup\text{AU}$) and (ii) uncertain from confident errors (C2 vs.\ C3). Each contrast is judged against $H_{\mathrm{ent}}$ with paired bootstrap intervals; Appendix~\ref{app:rq2} gives the rules.

H2b (steering) moves every question along $\dmu$ toward ``unknown'' at $\kappa_0$.
(i) Abstention should rise more on C2 than on C3, (ii) fewer than 10\% of C1 answers should become correct, and (iii) the probe end of the family should produce less abstention than $\dmu$ at the same $K$. H2b(i) holds if the bootstrap 95\% interval of the difference between the abstention gains on C2 and C3 lies above zero, H2b(ii) is judged by the point estimate, and H2b(iii) holds if the interval of the difference in abstention over all questions between the probe end and $\dmu$ lies below zero.

H2c (probe, then steer) steers only the questions the probe flags; it should keep at least 80\% of the abstention gain of H2b on C2 with at most half of its over-abstention on C0, judged by point estimates at the strength of $\dmu$ (Appendix~\ref{app:gate} varies the gate threshold).
Every direction is compared with a random direction at the same $K$.
Across models, we also measure, without a hypothesis, whether the models agree on which questions have a reachable answer and whether their probes and directions for ``unknown'' transfer.
\section{Experiments}
\label{sec:experiments}

\subsection{Models and Datasets}
\label{sec:models-data}

We study LLaMA-3.1-8B-Instruct, Qwen3.5-9B (the post-trained checkpoint), and Gemma-3-12B-it, with raw prompts and no chat template. We abbreviate these as LLaMA, Qwen, and Gemma, and in tables as L, Q, and G; figure labels omit the -Instruct and -it suffixes.
For each model, we use a layer at about $0.44$ of its depth, layer 14 of 32, 14 of 32, and 21 of 48, respectively (layers counted from 0).
We chose this depth in advance. A sweep over ten depths in all three models and for all three concepts (Appendix~\ref{app:layers}) supports it: averaged over the concepts, the probing AUROC at this layer is within 0.004 of the best layer in every model, and H1a holds at 59 of the 60 combinations of layer, model, and concept for truthfulness and sentiment, the two concepts with a steering sweep.

For \emph{truthfulness}, we use TruthfulQA \citep{lin2022truthfulqa}, each question followed by the answers of its multi-true list (MC2), labeled true or false (Appendix~\ref{app:data}); for \emph{sentiment}, we use 2,000 Yelp reviews (1,000 positive and 1,000 negative) \citep{zhang2015character-yelp-polarity}.
Both datasets are split 50/25/25 into an estimation subset and two held-out sets, on which the directions are evaluated.
For RQ2, we use 4,000 PopQA questions \citep{mallen2023popqa}; the concept ``unknown'' is used only in this experiment, with labels taken from each model's own answers (Section~\ref{sec:eval-know}).
Neutral WikiText-2 sentences \citep{merity2017pointer} measure the side effect $K$ and serve to fit the maps between models.

\subsection{Probing and Steering Evaluation}
\label{sec:eval}

For each model and concept, we estimate $\mu_0$, $\mu_1$ and $\SigW$ on the estimation subset, with the pooled estimate $\SigW=\frac{1}{n-2}\sum_{c\in\{0,1\}}\sum_{i:\,y_i=c}(h_i-\hat\mu_c)(h_i-\hat\mu_c)^\top$, which weights the two class covariances by their sample shares as in Section~\ref{sec:directions}, and compute $w_\lambda$ for 13 values of $\lambda/\tau$, spaced by half a decade from $10^{-3}$ to $10^{3}$, adding $\dmu$ as the end point.
We measure $R(\lambda)$ on both held-out splits (2,938 TruthfulQA answers; 1,000 reviews).

The steering effect $S(\lambda)$ is label-dependent: the change in the margin between the true answer and the best false one on TruthfulQA's single-true answer list (MC1) (Appendix~\ref{app:data}) or, for sentiment, the change in the log-odds of positive versus negative words, measured on the held-out questions or reviews. The side effect $K$, by contrast, is label-free and sets only the budget at which directions are compared.

We steer at six strengths, $a\in\{0.02,0.05,0.1,0.2,0.4,0.8\}$, and read off the effect at $\kappa_0$ by interpolating it linearly in $\sqrt{K}$ between the two consecutive strengths whose side effects bracket $\kappa_0$, with $K$ computed on 200 neutral WikiText-2 sentences. 
The budget $\kappa_0$ is the value of $K$ produced by $\dmu$ at $a=0.1$ for LLaMA and Qwen and at $a=0.01$ for Gemma, whose residual norm at this layer is far larger, separately for each model and concept; Gemma's strengths are all scaled by the same factor.
At a quarter to four times this budget, the best steering direction still lies at $3\tau$ or beyond in every combination (Appendix~\ref{app:measures}).
Intervals come from a paired bootstrap over held-out items (1,000 resamples; for TruthfulQA, a question is resampled together with all its answers), which also gives the plateaus of H1a; pairs in H1c are classified by point estimates, as fixed in advance.

\subsection{Cross-Model Transfer}
\label{sec:eval-transfer}

For each of the six ordered model pairs, we fit the maps of Section~\ref{sec:transfer} on 12,800 neutral texts (6,400 WikiText-2 sentences, each as its first half and in full), choose the ridge strength on 3,200 held-out texts, and report the held-out $R^2$ on neutral text and on the concept texts.
Probe transfer evaluates $M_{B\rightarrow A}^{\top}w_{A,\lambda}$ as a probe in $B$ on $B$'s held-out examples, against $B$'s own $w_{B,\lambda}$ with a paired bootstrap.
Steering transfer steers $B$ with $M_{A\rightarrow B}w_{A,\lambda}$ for $\lambda\in\{\tau,10\tau\}$ and with $M_{A\rightarrow B}\dmu_A$, each read off at $B$'s own $\kappa_0$.

\subsection{Answer-Reachability Experiment}
\label{sec:eval-know}

Each answer is judged by matching it against PopQA's gold answer and its aliases (Appendix~\ref{app:data}).
The six answers give the label of ``unknown'', $y=1$ if none of them is correct (``no correct answer'') and $y=0$ if at least one is (``some correct answer'').
Unlike the labels of TruthfulQA and Yelp, these labels depend on the model. Because the models abstain on most questions, the positive class is dominated by abstentions: AU makes up 82\%, 92\% and 68\% of it in LLaMA, Qwen and Gemma, wrong answers the rest, so $\dmu$ mainly separates questions the model will abstain on from questions it will answer correctly; a direction estimated from wrong answers alone was not run.
The probe is cross-fitted over two folds of the questions, with $\lambda$ chosen inside each training fold.
For every contrast of H2a, we compare it with $H_{\mathrm{ent}}$ and with a combination of both (Appendix~\ref{app:rq2}).

For steering, the directions of each fold are computed from the other fold; no question is steered with a direction that has seen it.
We compare five directions: $\dmu$ (mean of ``no correct answer'' minus mean of ``some correct answer''), the probe end $w_{\lambda^*}$, the same $\dmu$ applied only to the questions the probe flags (gated), a random unit direction, and the $\dmu$ of each other model carried over by the steering law.
Each is applied from the last prompt token onward at four strengths, $a\in\{0.05,0.1,0.2,0.4\}$ (ten times smaller for Gemma), and the greedy answer (at most 24 new tokens) is judged again.

We report each change in percentage points (pp), the percentage of the questions in a case that end with a given outcome (an abstention, or a correct answer) after steering minus the same percentage in an unsteered rerun.
The budget $\kappa_0$ is the $K$ that the ``unknown'' $\dmu$ produces at $a=0.1$ ($0.01$ for Gemma); Appendix~\ref{app:rq2} gives its values and how each direction is read at it.
Intervals come from a bootstrap over questions within each case (2,000 resamples).
We report 95\% intervals throughout and do not correct for multiple comparisons, except in the pooling ablation (Appendix~\ref{app:pool}), which uses Holm's method.

\section{Results}
\label{sec:results}

\subsection{RQ1: probing versus steering}
\label{sec:res-rq1}

\begin{figure}[t]
\centering
\includegraphics[width=0.9\linewidth]{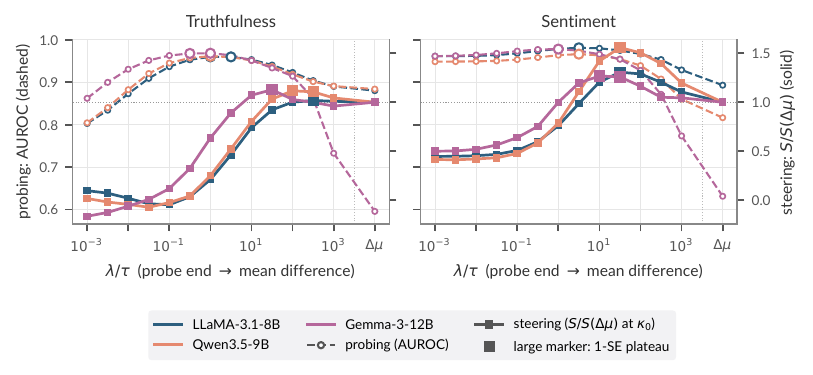}
\caption{Probing (dashed, held-out AUROC, left axis) and steering relative to $\dmu$ (solid, right axis) along the family; large markers mark the one-standard-error plateaus. Probing is best near $\lambda=\tau$, steering further toward $\dmu$.}
\label{fig:rq1-dial}
\vspace{-5pt}
\end{figure}
For H1a, Figure~\ref{fig:rq1-dial} plots probing and steering along the family. In all six combinations of model and concept, probing works best near $\lambda=\tau$, where the added term $\lambda I$ equals the average within-class variance.
There, the held-out AUROC is 0.96--0.98, indicating that a random example with the concept scores higher than a random example without it 96--98\% of the time.
It drops toward the mean difference (0.60--0.89 at $\dmu$).
Steering shows the opposite trend: the strongest directions lie between $10\tau$ and $316\tau$.
The steering plateau therefore lies to the right of the probing plateau in all six cases, as H1a predicts. A linear prediction in which only the component of the push along $\dmu$ moves the behavior, predicts the position of the steering plateau in all six cases (Appendix~\ref{app:firstorder}).
Because the probe and steering texts differ, we also ran a matched control (Appendix~\ref{app:matched}): H1a still holds in all six combinations, while the sentiment gain of H1b disappears in LLaMA and Gemma and shrinks in Qwen.

\begin{wrapfigure}{r}{0.47\linewidth}
\vspace{-5pt}
\centering
\includegraphics[width=\linewidth,trim=2 2 2 14,clip]{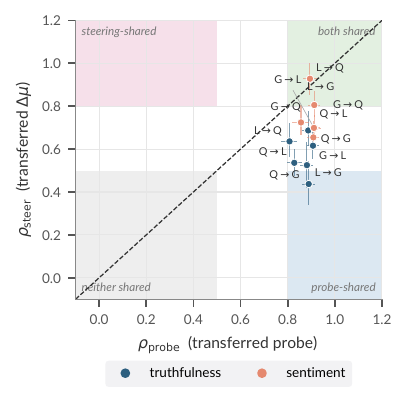}
\vspace{-5pt}
\caption{H1c: $\rho_{\mathrm{probe}}$ (horizontal) and $\rho_{\mathrm{steer}}$ (vertical) per ordered pair and concept. Most pairs lie below the diagonal.}
\label{fig:xmodel}
\vspace{-10pt}
\end{wrapfigure}
In our main setting, the separation between the two optima depends on the concept.
For truthfulness, the best probe is a weak steering direction: at $\lambda=\tau$ it changes the answer margin only 21--64\% as much as $\dmu$ does at the same side effect.
No direction on the path steers clearly better than $\dmu$; the largest gain, 9.6\% in Qwen, falls just short of the 10\% we required in advance, and H1b fails in all three models.
For sentiment, the best probe (at $3\tau$ in LLaMA and Qwen, at $\tau$ in Gemma) already steers about as well as $\dmu$ (0.99--1.11 of its effect), and here H1b holds in all three models: a $\lambda^*$ chosen on one held-out split gives 27--56\% more effect than $\dmu$ on the other.
The activation-pattern argument of Section~\ref{sec:directions} (Eq.~\eqref{eq:pattern}) predicts the truthfulness case, where steering is best at $\dmu$. The apparent exception for sentiment is largely attributable to the different probe and steering texts (Appendix \ref{app:matched}).

\paragraph{Across models.}
For H1c, we carry probes and steering directions from each model into every other model and compare them with the target's own through the two ratios of Eq.~\eqref{eq:rho} (Figure~\ref{fig:xmodel}; values in Figure~\ref{fig:app-matrices}, Appendix~\ref{app:figs}).
A carried probe keeps 81--91\% of the target probe's above-chance AUROC in all twelve pairs, while the carried $\dmu$ reaches 44--93\% of the target's own steering effect, less than $\rho_{\mathrm{probe}}$ in 11 of 12 pairs.

By the categories fixed in advance for H1c, nine pairs are inconclusive, one is probe-shared, and two sentiment pairs (LLaMA$\rightarrow$Qwen and Gemma$\rightarrow$Qwen) are both-shared, mostly because $\rho_{\mathrm{steer}}$ falls between 0.5 and 0.8.
We therefore read the result as a descriptive tendency in the three tested models rather than a clean classification, in which probes transfer more reliably than the mean difference used for steering, each under its transfer law.
Appendix~\ref{app:figs-xmodel} discusses three caveats, among them that the maps fit the concept texts less well than neutral text ($R^2$ of 0.13--0.47 against 0.48--0.70).

Because the target probes lie near the AUROC ceiling, $\rho_{\mathrm{probe}}$ is compressed toward 1; on the separation scale $d'=\sqrt{2}\,\Phi^{-1}(\mathrm{AUROC})$, where $\Phi$ is the standard normal distribution function, the carried probe keeps 65--79\% of the target's $d'$, which still exceeds $\rho_{\mathrm{steer}}$ in 9 of 12 pairs, in all six for truthfulness, and in all six pairs for ``unknown'' (82--92\% against 30--79\%; Section~\ref{sec:res-rq2}).

\subsection{RQ2: answer reachability and abstention}
\label{sec:res-rq2}

The three models abstain on most of the 4,000 PopQA questions (57--78\%) and answer 11--21\% correctly.
Knowledge-available errors are rare (C1: 71, 33, and 35 questions for LLaMA, Qwen, and Gemma), probably because of a sampling bias of the labels: a question whose correct answer has probability $p$ under sampling is missed by all five samples with probability $(1-p)^5$, 0.77 at $p=0.05$ and 0.44 at $p=0.15$, and is then counted among the questions without a correct answer rather than in C1.
C1 is therefore a conservative set with wide intervals, the 10\% limit of H2b(ii) amounts to 7, 3 and 3 questions, and C2 contains some questions that more samples would have moved to C1; C0 is defined by the greedy answer alone and is unaffected.
The C2-versus-C3 contrasts are not affected in the same way: C2 holds 180, 123 and 328 questions and C3 259, 83 and 576, so even the smallest class, Qwen's C3, is more than twice the largest C1.

For H2a(i), the probe separates questions with and without a reachable answer better than the first-token entropy does, which supports H2a(i) among wrong answers in LLaMA and Gemma and, once abstentions are included, in all three models.
For H2a(ii), the probe flags uncertain errors more often than confident ones in all three models (by 21 to 32\,pp), which meets the flag-rate rule, but it does not separate the two better than that entropy in any model (Figure~\ref{fig:app-rq2-read}).

\begin{figure}[t]
\centering
\includegraphics[trim=0 1cm 0 0, clip, width=1\linewidth]
{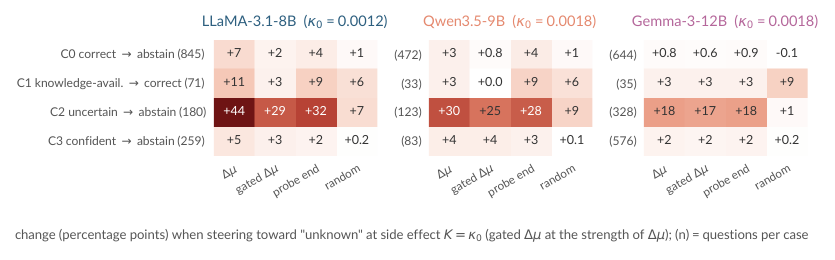}
\caption{Change (percentage points) of C0--C3 under steering toward ``unknown'' (H2b, H2c), with the questions per case in brackets. $\dmu$ turns far more uncertain (C2) than confident (C3) errors into abstentions; a random direction at the same side effect also moves C2 more than C3, but by far less.}
\label{fig:rq2-cases}
\vspace{-10pt}
\end{figure}

For H2b, Figure~\ref{fig:rq2-cases} shows how each case changes under four of the directions of Section~\ref{sec:eval-know} at $K=\kappa_0$, the gated $\dmu$ at the strength of $\dmu$.
Steering along $\dmu$ turns many uncertain errors into abstentions (C2: +44, +30 and +18\,pp for LLaMA, Qwen and Gemma) but few confident errors (C3: +5, +4 and +2), at a small cost on C0 (+7, +3 and +1); H2b(i) holds in all three models.
A random direction with the same $K$ raises abstention on C2 by only 7, 9 and 1\,pp.
Pushed harder, it also moves C2 far more than C3 (Figure~\ref{fig:app-rq2-grid}), so uncertain answers give way to any push, and what $\dmu$ adds is more abstentions at the same side effect.

For H2b(ii), some knowledge-available errors do become correct (11, 3 and 3\%), but a random direction produces about as many (6, 6 and 9\%). Only in LLaMA does this exceed the 10\% limit we set in advance; H2b(ii) fails there and holds in Qwen and Gemma. With 33--71 C1 questions, these percentages have wide intervals, so the H2b(ii) verdicts rest on point estimates, as the rule fixed in advance requires.
We read these changes as small pushes on answers that were nearly tied, not as retrieval of the fact.

For H2b(iii), the probe end abstains less than $\dmu$ only in LLaMA (C2: +32 against +44), over all questions $-2$ pp, interval $[-2,-1]$) as predicted; in Qwen and Gemma it abstains about as much (+28 against +30, +18 against +18).

For H2c, gating passes the test we fixed in advance only in Qwen, and the verdicts are the same for thresholds between the 85th and 95th percentile of the C0 scores (Appendix~\ref{app:gate}).
At the strength of $\dmu$, gating lowers the cost on correct answers in LLaMA and Qwen (C0: +2 instead of +7, and +0.8 instead of +3) and keeps 65 and 84\% of the gain on uncertain errors; in Gemma it keeps 95\% of the gain but saves little of the cost (+0.6 instead of +0.8).

Two further comparisons qualify this (Appendix~\ref{app:gate}).
Gating by the first-token entropy instead of the probe, at the same 90th-percentile threshold, passes the rule in LLaMA (90\% of the gain at 46\% of the cost) and fails in Qwen and Gemma, so the two gates pass in different models and neither is uniformly better.
Compared with ungated $\dmu$ at a strength that costs the same on C0, the probe gate keeps more of the gain on uncertain errors in all three models (29 against 25\,pp in LLaMA, 25 against 12 in Qwen, 17 against 15 in Gemma).
In Gemma, the cost ratio of 0.80 rests on 4 against 5 correct answers lost out of 644, so its verdict is a matter of one question.

%%%%%It is only in RQ2 and answers no hypothesis. H2c compares gating at the same strength. The comparison at the same gain was added afterwards and decides nothing:

%Compared at the same gain instead of the same strength, an unplanned comparison, gating lowers the cost in all three models, by 34 to 59\% at the strongest push (Figure~\ref{fig:app-rq2-plane}).

Across models, the tendency of RQ1 also appears for ``unknown''.
A probe carried into another model keeps 89 to 96\% of the target probe's above-chance AUROC, while a carried $\dmu$ reaches 30 to 79\% of the target's own abstention gain on uncertain errors, and the models agree only moderately on which questions they cannot answer (Cohen's $\kappa$ of 0.56 to 0.65; Table~\ref{tab:rq2-xmodel} in Appendix~\ref{app:rq2}).

% decided not to: add gating with the partner's probe (xgate, qmatch) and per-question patching once Qwen's runs finish.

\section{Related Work}
\label{sec:related}

\paragraph{Probing and steering concepts.}
A unified analysis of steering methods favors the mean of differences \citep{im2025unified}, and ridge-regularized probes have been used to obtain concept directions for steering \citep{gao2026raptor}.
Closest to our setting, \citet{cheng2026liseco} and \citet{hedstrom2025steer} steer along a probe's own direction with a strength set per token in closed form, in several model families; the first notes that linear decodability does not guarantee linear controllability, which agrees with our finding that the best probe is not the best steering direction.
These works vary the strength along one direction; none of them, nor the comparisons cited in Section~\ref{sec:intro}, varies the direction itself along the family between the probe and the mean difference and measures probing and steering at each point of it at a fixed side effect.
Concept erasure removes several directions \citep{ravfogel2020null} or, in closed form, the subspace that linear classifiers use \citep{belrose2023leace}; we study single directions.
%\paragraph{Representation transfer.}
%Learned maps can transfer features, probes, and interventions between models \citep{chen2025stitching,bello2025lrt,oozeer2025activation,gorbett2026linear}.
%We compare how probes and steering directions transfer across families, each by its own law.

\paragraph{Representation transfer.}
Learned maps carry objects from one model to another: affine stitches transfer sparse autoencoders, sparse probes and steering vectors between sizes of one family \citep{chen2025stitching}, affine maps and autoencoders transfer steering vectors between sizes and across families \citep{bello2025lrt,oozeer2025activation}, and an affine map on final hidden states lets one model's states be decoded through another's output head \citep{gorbett2026linear}.
These works move one kind of object at a time, mostly steering vectors, and judge it by the behavior it produces in the target; where probes and steering vectors are both moved \citep{chen2025stitching}, the two are tested on different pairs of models, so their transfer cannot be compared.
We instead move the same family of directions in both roles between the same three model families, in all six ordered pairs, each role under its own law and both measured against the target's own direction at a fixed side effect, with the wrong law and an orthogonal map as controls; this shows that probes transfer more reliably than steering directions and that each role transfers well only under its own law.

\paragraph{Knowledge awareness.}
Whether a model knows an answer can be predicted from its own judgment \citep{kadavath2022language} or its internal states \citep{gottesman2024keen}, and errors made despite knowledge can be distinguished from errors caused by missing knowledge \citep{simhi2024distinguishing,orgad2025llms}.
A probe on the hidden state before generation can also stand in for sampling-based semantic uncertainty \citep{han2024sep}.
Other studies examine confident errors and association-driven hallucinations \citep{simhi2025trust,cheang2025llms}, steer knowledge awareness or gate the intervention \citep{ferrando2025know,kumaran2026causal,hedstrom2025steer,lee2025cast}, or train steering toward abstention into the weights of one layer \citep{yang2026casal}.
We instead study what steering toward ``unknown'' does to each type of error, and how its probe and steering direction differ and transfer.
As an application, this steering reduces hallucinations by turning wrong answers into abstentions, mostly the uncertain ones, at a small cost in correct answers.

\section{Conclusion}

In each model we test, the direction of the family that probes a concept best does not steer it best: probing peaks near $\lambda=\tau$ and steering further toward the mean difference, in all six main combinations, also with matched texts, at 59 of 60 combinations for truthfulness and sentiment in a layer sweep, and in the four cases where rotation reaches the same side effect.
The two roles also separate across models.
Each carried under its own law, probes keep most of their performance, for ``unknown'' as well, while the mean difference keeps less of its steering effect in most pairs; under maps fitted on neutral text, the tested models agree more on how a concept is read than on how it is changed, although part of the steering loss may come from the maps.
Where steering works best is less stable: whether an intermediate direction beats the mean difference depends on the concept, the layer, the operator, and the texts (Appendices~\ref{app:matched}, \ref{app:rotation} and~\ref{app:layers}).
On PopQA, steering toward ``unknown'' turns uncertain errors into abstentions far more readily than confident ones; the probe for ``unknown'' steers almost as strongly as the mean difference in Qwen and Gemma, and gating by the probe protects correct answers but passes our test only in Qwen.

These results come from three concepts and three models, with the main analyses at one layer per model; the twelve ordered pairs of the cross-model experiments share models and are therefore not independent, and all results concern single directions under additive and rotational steering. Appendix~\ref{app:limits} lists the further limitations.
Probe directions should therefore not be reused for steering without checking how well they steer.

\section*{Author Contributions}
Afshin Sadeghi conceived the study, developed the conceptual framework and the hypotheses, designed and ran the experiments, analyzed the results, and wrote the paper. Daniel Becker provided feedback on the concept of the family of directions. This work was carried out in the AI Assurance and Assessments department of the Fraunhofer (IAIS) Institute.

\bibliographystyle{plainnat}
\bibliography{references}

\clearpage % or \cleardoublepage
\appendix
\noindent{\Large\bfseries Appendix}\par
\vspace{0.3em}

\section{Limitations}
\label{app:limits}

\paragraph{Scope of the evidence.}
All results come from three concepts (truthfulness, sentiment, and ``unknown'' on PopQA) and three instruction-tuned models (LLaMA-3.1-8B, Qwen3.5-9B, Gemma-3-12B), and the main analyses use one layer per model.
The layer sweep of Appendix~\ref{app:layers} covers the probing and steering of RQ1 only; the steering of RQ2, including the H2b(iii) and H2c verdicts, was run at the paper's layer alone, so we do not know how those verdicts change with depth.
Other concepts, base models, or larger models may behave differently.

\paragraph{The cross-model pairs are not independent.}
Three models give six ordered pairs, and two concepts give twelve.
Every pair shares its source or its target with four other pairs, so a property of one model appears in four of the twelve points at once; for example, the poor probing of Gemma's own mean difference affects every pair into or out of Gemma.
Counts such as ``11 of 12 pairs'' therefore describe a pattern in three models and should not be read as twelve separate confirmations.
The cross-model comparison also uses only the unmatched probe and steering texts of Section~\ref{sec:eval}; the matched-text control of Appendix~\ref{app:matched} was run within models only.

\paragraph{Single directions and two operators.}
We study one direction at a time, drawn from a single one-parameter family, and two ways of applying it: addition and a gated rotation.
We do not study subspaces, nonlinear probes, sparse-autoencoder features, or steering operators learned from data, and the conclusions may not carry over to them.
The rotation operator is gated by the state's cosine with the direction, and its side-effect budget is set on neutral text where the gate opens less often than on task text; its H1b verdicts therefore mix the effect of the rotation with that of the gate (Appendix~\ref{app:rotation}).

\paragraph{What is measured.}
Steering is measured by the change of an answer margin (truthfulness, ``unknown'') or of word log-odds (sentiment), not by task accuracy, so a direction that moves the margin need not change which answer the model gives.
Probing is measured by AUROC.
The two transfer ratios $\rho_{\mathrm{probe}}$ and $\rho_{\mathrm{steer}}$ therefore compare different quantities, an AUROC above chance and a steering effect, and their comparison is approximate.
The side effect $K$ is the next-token KL divergence on 200 neutral WikiText-2 sentences, not on the task prompts or over generated continuations, and the budget $\kappa_0$ is anchored to the mean difference at one strength; the post hoc check over a sixteen-fold range of budgets (Appendix~\ref{app:measures}) reads saved curves and has no intervals.

\paragraph{Maps and estimates.}
The maps between models are fitted on neutral text and fit the concept texts less well ($R^2$ of 0.13--0.47 against 0.48--0.70 on held-out neutral text), so part of the loss of a carried steering direction may come from the map rather than from the models, and no pair can be called ``neither shared''.
$\SigW$ is estimated once per model and concept from fewer examples than dimensions, so the estimate is singular and the small-$\lambda$ end of the family is the part of $\Delta\mu$ outside the span of the training states rather than the Fisher discriminant (Appendix~\ref{app:witness}).

\paragraph{RQ2 labels.}
The cases of RQ2 are defined from the greedy answer and five sampled answers, so they capture whether a correct answer is reachable under this sampling, not whether the model knows the fact; a question whose correct answer appears with probability $p$ in one sample is missed by all five with probability $(1-p)^5$, which biases the ``no correct answer'' class toward questions the model could sometimes answer.
Knowledge-available errors (C1) are rare, 71, 33, and 35 questions, so the H2b(ii) verdicts rest on point estimates with wide intervals.
Prompts are plain text without the models' chat templates, and Qwen's thinking tokens are handled by a token budget (Appendix~\ref{app:rq2}).

\paragraph{Truthfulness data.}
Truthfulness is tested on TruthfulQA alone, and its directions are computed and evaluated on the same dataset, so a gain of an intermediate direction cannot be separated from a gain of fitting the test distribution.
Truthfulness probes trained on one dataset can generalize poorly to others \citep{liu2024universal}, and we did not test whether the directions of this paper transfer to other truthfulness data.

\section{The two ends of the family as divergence witnesses}
\label{app:witness}

Under a shared-covariance Gaussian model, the score of $\wlda$ is a monotone function of the likelihood ratio, so no other score separates the classes with a higher area under the ROC curve.

In our setting $n<d$ and $\SigW$ is singular.
The small-$\lambda$ end therefore approaches the component of $\dmu$ outside the span of the within-class deviations.
Let $V$ hold the orthonormal eigenvectors of $\SigW$ with positive eigenvalues $e_k>0$, obtained from the thin SVD of the centered data; its columns span the within-class deviations.
Then
\[
\lambda(\SigW+\lambda I)^{-1}\dmu
=
V\operatorname{diag}\!\left(\frac{\lambda}{e_k+\lambda}\right)V^\top\dmu
+
(I-VV^\top)\dmu,
\]
which tends to $(I-VV^\top)\dmu$ as $\lambda\rightarrow0$.
Since
$\SigW+\lambda I\propto(1-\gamma)\SigW+\gamma\tau I$
with $\gamma=\lambda/(\lambda+\tau)$, $w_\lambda$ is linear regularized discriminant analysis \citep{friedman1989rda}, which shrinks $\SigW$ toward the spherical covariance $\tau I$ with the same total variance.
At $\lambda=\tau$ both get equal weight ($\gamma=\tfrac12$); at $10^{-3}\tau$ almost all weight is on $\SigW$, and at $10^{3}\tau$ almost all on $\tau I$.
The point $\lambda=\tau$ is thus defined by this isotropic shrinkage target, not as a general balance between probing and steering; another covariance estimator would place the family's points differently.

The probe end also stands for the probes used in practice.
Linear probes are a standard tool for reading concepts from hidden states \citep{belinkov2022probing} and are most often trained by logistic regression \citep{li2023iti,orgad2025llms,kossen2024sep}; for Gaussian classes with a shared covariance, the population direction of logistic regression is the covariance-adjusted mean difference, a standard property of the logistic model.
Covariance-adjusted mean differences are also used directly as probes \citep{marks2024geometry,lee2026meanshift}, and an estimator in the style of linear discriminant analysis gives the concept vectors of \citet{park2025categorical}.

The two endpoints can be viewed as witnesses of different divergences between the class distributions $P_1$ and $P_0$ of $h$.
For a linear kernel,
\[
\sup_{\|w\|\le1}
\mathbb{E}_{P_1}[w^\top h]-\mathbb{E}_{P_0}[w^\top h]
=\|\dmu\|,
\]
attained at $w=\dmu/\|\dmu\|$ \citep{gretton2012kernel}.
The variational witness of an $f$-divergence is
$f'(dP_1/dP_0)$ \citep{nguyen2010estimating}.
Under the shared-covariance Gaussian model, this is a monotone function of $\wlda^\top h$.
\citet{ribero2026regularized} estimate $f$-divergence witnesses with a regularized kernel likelihood-ratio estimator.
Our operator is different because it is built from $\SigW$, so we do not claim that the two estimators coincide.

\section{Transfer laws and maps}
\label{app:derive}
If the state of $A$ moves from $h_A$ to $h_A+\delta$, the mapped state of $B$ moves by $M_{A\rightarrow B}(h_A+\delta)+b_{A\rightarrow B}-(M_{A\rightarrow B}h_A+b_{A\rightarrow B})=M_{A\rightarrow B}\delta$, since the offset cancels; this gives the steering law.
Writing the state of $A$ through the reverse map, $h_A\approx M_{B\rightarrow A}h_B+b_{B\rightarrow A}$, gives $w_A^\top h_A\approx(M_{B\rightarrow A}^{\top}w_A)^\top h_B+w_A^\top b_{B\rightarrow A}$, which gives the probe law; the constant only shifts the score, which does not change the AUROC.

We fit
\[
(M_{A\rightarrow B},b_{A\rightarrow B})
=
\arg\min_{M,b}
\sum_x\|h_B(x)-Mh_A(x)-b\|^2+\alpha\|M\|_F^2,
\]
where the sum runs over the neutral training texts $x$, $\|\cdot\|_F$ is the Frobenius norm, and the ridge strength $\alpha$ is selected on held-out neutral sentences.
$M_{B\rightarrow A}$ and $b_{B\rightarrow A}$ are fitted in the same way with $A$ and $B$ exchanged.
The neutral set does not overlap the concept or KL sets.
For the orthogonal Procrustes control, we use the partial isometry $Q=U_rV_r^\top\in\mathbb{R}^{d_B\times d_A}$, where $U_rS_rV_r^\top$ is the thin singular value decomposition, of rank $r=\min(d_A,d_B)$, of the cross-covariance $\sum_x(h_B(x)-\bar h_B)(h_A(x)-\bar h_A)^\top$ of the neutral states and $\bar h_A$, $\bar h_B$ are their means; $Q$ has orthonormal columns if $d_A\le d_B$ and orthonormal rows otherwise.
Using $Q$ as $M_{A\rightarrow B}$ and $Q^\top$ as $M_{B\rightarrow A}$ makes the two transfer laws coincide, since $(Q^\top)^\top w=Qw$.
$Q$ carries no scale, but we use it only for probing and for cosines, after normalizing the carried direction to unit length, so the difference in hidden-state scale between models does not enter.
We also evaluate the wrong transfer law, a label-permutation null, and held-out $R^2$ (Appendix~\ref{app:figs-xmodel}, Table~\ref{tab:xmodel-controls}).

\section{Side effect and steering measures}
\label{app:measures}
For each intervention, we use $K$ as defined in Eq.~\eqref{eq:K}.
For small changes, smooth $f$-divergences agree to second order up to a constant, so KL fixes the units of the side-effect budget.
For a small push, $K$ grows with the square of the strength, so $\sqrt K$ grows approximately in proportion to it; effects are therefore interpolated linearly in $\sqrt K$ between consecutive strengths, in RQ1 and RQ2 alike.
With $(K_0,S_0)=(0,0)$ for the unsteered model and $(K_j,S_j)$ the side effect and effect at the $j$-th strength, and $j$ such that $K_j\le\kappa_0\le K_{j+1}$,
\[
S(\kappa_0)=S_j+(S_{j+1}-S_j)\,\frac{\sqrt{\kappa_0}-\sqrt{K_j}}{\sqrt{K_{j+1}}-\sqrt{K_j}} ;
\]
the origin is thus one node of the interpolation, and in every RQ1 run $\kappa_0$ falls between the third and fourth strength, so the two neighbouring measured points determine the value.

The truthfulness margin is
$m(q)=L(c^*)-\max_{c\ne c^*}L(c)$, with answer-only log-likelihoods $L$.
For sentiment, we use
$\log\sum_{v\in V^+}p(v\mid x)-\log\sum_{v\in V^-}p(v\mid x)$
after ``Overall, the experience was''.
For Gemma we use $a=0.01$ instead of $0.1$, because its residual norm at the layer we use is about $4\cdot10^{4}$ (against about 10 and 40 for LLaMA and Qwen) and $a=0.1$ already gives a side effect about 100 times larger.
$\kappa_0$ is then of the same order in all three models, 0.0013, 0.0020, and 0.0020 nats for truthfulness and 0.0016, 0.0024, and 0.0028 nats for sentiment (LLaMA, Qwen, and Gemma).
In every RQ1 run, $\kappa_0$ lies between the side effects of two measured strengths of each direction, the third or fourth of six, and the largest strength gives a side effect at least 27 times $\kappa_0$.
The budget is measured at the next token of neutral sentences, not on the task prompts or over generated continuations.

The budget is anchored to $\dmu$.
Since each effect is interpolated in $\sqrt K$, the anchor sets only the level of side effect at which all directions are compared; in a post hoc check we read the saved effect curves again at $0.25$, $0.5$, $2$ and $4$ times $\kappa_0$ (mean effects, no new runs and no intervals).
Over this sixteen-fold range, the direction with the largest steering effect lies at $3\tau$ or beyond in all six combinations of model and concept, and at $\dmu$ itself only once (truthfulness in Qwen at $4\kappa_0$).
At $\lambda=\tau$, near the best probe, the truthfulness effect stays at 13--27\% of that of $\dmu$ in LLaMA and Qwen and at 61--72\% in Gemma, and the sentiment effect at 76--105\%.

The full run took about 24 hours of job time, each job on two 80\,GB GPUs; the cross-model maps and the PopQA analysis run on CPU.

\section{A linear prediction of the steering curve}
\label{app:firstorder}

If the behavior responds, to first order, only to the component of the push along $\dmu$, then adding $a\,w_\lambda$ changes the margin by $a\cos(w_\lambda,\dmu)\,g$ for a constant $g$, and at a fixed side effect
\[
\frac{S(\lambda)}{S(\dmu)}\approx\frac{a_\lambda(\kappa_0)}{a_{\dmu}(\kappa_0)}\cos(w_\lambda,\dmu),
\]
where $a_\lambda(\kappa_0)$ is the strength at which $w_\lambda$ reaches the budget, read from the measured curve of $K$ against $a$ as in Appendix~\ref{app:measures}.
The prediction uses the cosines and the side-effect curves only; the one steering measurement it needs is $S(\dmu)$.
Table~\ref{tab:firstorder} compares it with the measured curve.
It reproduces the shape of the curve and the position of the steering plateau in all six cases (correlation 0.91--1.00 over the 14 directions) and the values in LLaMA and Qwen for truthfulness.
The probe end steers weakly because its cosine with $\dmu$ is 0.25--0.46 and the budget lets it push only 1.3--1.5 times as hard as $\dmu$.
It under-predicts in two places.
For sentiment in LLaMA and Qwen the measured gain of the intermediate directions exceeds the predicted one (1.29 and 1.51 against 1.12 and 1.19 at $100\tau$), so the component orthogonal to $\dmu$ also steers there.
In Gemma the prediction is low throughout, because Gemma's $\dmu$ carries a large component that neither probes (Appendix~\ref{app:figs}) nor steers, which lowers every cosine with it.
These two deviations are the cases that Section~\ref{sec:res-rq1} and Appendix~\ref{app:matched} report as unstable.

\begin{table}[h]
\centering
\small
\begin{tabular}{llcccc}
\hline
 & & \multicolumn{3}{c}{$S(\lambda)/S(\dmu)$, predicted / measured} & \\
Concept & Model & $\lambda=\tau$ & $10\tau$ & $100\tau$ & $r$ over the 14 directions\\
\hline
Truthfulness & LLaMA & 0.38 / 0.21 & 0.74 / 0.74 & 0.96 / 1.00 & 0.98\\
 & Qwen & 0.33 / 0.25 & 0.79 / 0.80 & 1.11 / 1.12 & 1.00\\
 & Gemma & 0.33 / 0.64 & 0.53 / 1.07 & 0.58 / 1.03 & 0.91\\
\hline
Sentiment & LLaMA & 0.59 / 0.76 & 0.93 / 1.21 & 1.12 / 1.29 & 0.97\\
 & Qwen & 0.63 / 0.79 & 1.03 / 1.42 & 1.19 / 1.51 & 0.97\\
 & Gemma & 0.57 / 1.00 & 0.83 / 1.27 & 0.86 / 1.16 & 0.92\\
\hline
\end{tabular}
\caption{First-order prediction of the steering effect along the family, relative to $\dmu$, against the measured value at $K=\kappa_0$, and the Pearson correlation of the two curves over all 14 directions.}
\label{tab:firstorder}
\end{table}

\section{Matched probing and steering texts}
\label{app:matched}

In the main runs, the texts that the probe reads differ from the steering prompts.
For truthfulness, the probe reads ``Q: \{question\} A: \{answer\}'', and the steering prompt adds an instruction before it.
For sentiment, the probe reads the first 120 words of a review, and the steering prompt consists of its first 24 words followed by ``Overall, the experience was''.

In a matched control, truthfulness uses one template, ``Q: \{question\}\textbackslash nA: \{answer\}'', without the instruction for both operations; it separates question and answer by a newline rather than the space of the main runs, so the matched control differs from them in the instruction and in this separator, for probing and steering alike.
The sentiment probe reads the steering prompt itself, at its last token, which is the position that is steered.

Directions, splits, strengths, and the rule for $\kappa_0$ are those of Section~\ref{sec:eval}; since $\dmu$ changes with the texts, so does $\kappa_0$.

\begin{table}[h]
\centering
\footnotesize
\setlength{\tabcolsep}{2.5pt}
\begin{tabular}{lllcccccccc}
\hline
Concept & Model & Texts & \begin{tabular}[b]{@{}c@{}}Best probing\\AUROC\end{tabular} & \begin{tabular}[b]{@{}c@{}}Probing\\plateau\end{tabular}&\begin{tabular}[b]{@{}c@{}}Steering\\plateau\end{tabular}& $\kappa_0$ & $S(\dmu)$ & \begin{tabular}[b]{@{}c@{}} $S(\tau)/$ \\ $S(\dmu)$ \end{tabular} & \begin{tabular}[b]{@{}c@{}} $S_{(2)}(\lambda^*)/$ \\ $S_{(2)}(\dmu)$ \end{tabular} \\
\hline
Truthfulness & LLaMA & paper & 0.960 & $\tau$--$3\tau$ & $316\tau$ & 0.0013 & 0.33 & 0.21 & 1.01\\
 & & matched & 0.958 & $\tau$--$3\tau$ & $100\tau$--$316\tau$ & 0.0015 & 0.25 & 0.29 & 1.04\\
 & Qwen & paper & 0.961 & $\tau$ & $100\tau$--$316\tau$ & 0.0020 & 0.60 & 0.25 & 1.10\\
 & & matched & 0.962 & $\tau$--$3\tau$ & $100\tau$ & 0.0018 & 0.51 & 0.19 & 1.08\\
 & Gemma & paper & 0.968 & $0.3\tau$--$\tau$ & $32\tau$ & 0.0020 & 0.56 & 0.64 & 1.07\\
 & & matched & 0.968 & $0.3\tau$--$\tau$ & $3\tau$--$10\tau$ & 0.0026 & 0.19 & 1.30 & 1.87$^{*}$\\
\hline
Sentiment & LLaMA & paper & 0.981 & $3\tau$ & $32\tau$ & 0.0016 & 0.85 & 0.76 & 1.31$^{*}$\\
 & & matched & 0.964 & $32\tau$ & $100\tau$--$316\tau$ & 0.0008 & 1.04 & 0.22 & 1.03\\
 & Qwen & paper & 0.966 & $3\tau$ & $32\tau$ & 0.0024 & 0.59 & 0.79 & 1.56$^{*}$\\
 & & matched & 0.941 & $3\tau$--$10\tau$ & $32\tau$--$316\tau$ & 0.0014 & 0.35 & 0.34 & 1.31$^{*}$\\
 & Gemma & paper & 0.978 & $\tau$ & $10\tau$--$32\tau$ & 0.0028 & 0.73 & 1.00 & 1.27$^{*}$\\
 & & matched & 0.964 & $3\tau$ & $10\tau$--$\dmu$ & 0.0021 & 0.72 & 0.54 & 0.99\\
\hline
\end{tabular}
\caption{RQ1 with the paper's texts and with matched probe and steering texts ($^{*}$: H1b holds). $\kappa_0$ in nats; $S(\dmu)$ is the held-out effect of $\dmu$ at $\kappa_0$, in nats of answer margin for truthfulness and in log-odds for sentiment. H1a holds in every case.}
\label{tab:matched}
\end{table}
With matched texts, H1a holds in all six combinations of model and concept (Table~\ref{tab:matched}).
For truthfulness, the probing plateau stays at $0.3\tau$--$3\tau$ in all three models and the steering plateau lies far to its right.

In Gemma, removing the instruction cuts the effect of $\dmu$ on truthfulness from 0.56 to 0.19, while $\lambda^*$ reaches 0.39 on the second split; the H1b ratio of 1.87 therefore reflects mainly the weaker $\dmu$, not a stronger intermediate direction.
The matched control changes two things at once for truthfulness, the instruction and the separator, and for sentiment it also shortens the text from which $\dmu$ and $\SigW$ are estimated to 24 words, which lowers the probing AUROC by 0.014--0.025; a control that adds the instruction to the probe text, or the closing phrase to the full review, would separate these effects and was not run.

For sentiment, the direction at $\lambda=\tau$ steers only 22--54\% as strongly as $\dmu$, instead of 76--100\%, and the gain of an intermediate direction disappears in LLaMA and Gemma (H1b ratios 1.03 and 0.99) and shrinks in Qwen (1.31 instead of 1.56).
The sentiment gain of H1b in Section~\ref{sec:res-rq1} therefore depends largely on reading the probe on the full review rather than on the steering prompt.
The robust result is that probing and steering favor different regions of the family. The stronger H1b claim is less stable under changes to the experimental setting (Appendices ~\ref{app:rotation} and ~\ref{app:layers}).

The robust result is that probing and steering favor different regions of the family. The stronger H1b claim is less stable under changes to the experimental setting (Appendices~\ref{app:rotation} and~\ref{app:layers}).
The matched control was run within models only; the cross-model comparison of H1c uses the unmatched texts of Section~\ref{sec:eval}, the rows marked ``paper'' in Table~\ref{tab:matched}.
In the matched truthfulness control the directions are still estimated at the last token of the answer, while steering acts from the last prompt token onward, so a positional difference remains there.

\section{Steering by rotation}
\label{app:rotation}

Spherical Steering \citep{you2026spherical} rotates the hidden state toward a direction instead of adding the direction to it.
We repeat the RQ1 measurements with a gated rotation in this style: $h$ is rotated toward $w$ along the great circle by $g\theta$ degrees, at most by the angle between them and with its norm kept, where $g=\min\{1,\max\{0,(c-\beta)/(1-\beta)\}\}$ and $c=-\tanh(\kappa\cos(h,w))$.
This $c$ equals their vMF gate score, the probability of alignment with $-w$ minus that with $w$ under a two-class softmax; we fix $\beta=-0.15$ and $\kappa=20$.
Unlike their method, which rotates by a fraction of the angle to $w$, we rotate by the angle $g\theta$, so that the side effect can be matched to the additive budget $\kappa_0$.
The angles $\theta$ run from 1 to 40 degrees (0.1 to 4 for Gemma), and every effect is read at the additive budget $\kappa_0$ of Section~\ref{sec:eval}, so the two operators are compared at the same side effect.

\begin{table}[h]
\centering
\small
\begin{tabular}{llcccc}
\hline
Concept & Model & Probing plateau & Steering plateau & H1a & $S_{(2)}(\lambda^*)/S_{(2)}(\dmu)$\\
\hline
Truthfulness & LLaMA & $\tau$--$3\tau$ & $32\tau$ & holds & 1.73$^{*}$\\
 & Qwen & $\tau$ & $32\tau$ & holds & 1.40$^{*}$\\
 & Gemma & $0.3\tau$--$\tau$ & $3\tau$--$10\tau$ & holds & 1.17\\
\hline
Sentiment & LLaMA & $3\tau$ & -- & not defined & --\\
 & Qwen & $3\tau$ & -- & not defined & --\\
 & Gemma & $\tau$ & $3\tau$--$10\tau$ & holds & 1.62$^{*}$\\
\hline
\end{tabular}
\caption{RQ1 with the gated rotation, read at the additive budget $\kappa_0$ ($^{*}$: H1b holds; --: the rotation does not reach $\kappa_0$). Where defined, the steering plateau lies to the right of the probing plateau.}
\label{tab:rotation}
\end{table}

Where the comparison is defined, H1a holds under rotation as well (Table~\ref{tab:rotation}): the steering plateau lies between $3\tau$ and $32\tau$, to the right of the probing plateau.
For truthfulness, an intermediate direction now steers clearly better than $\dmu$ in LLaMA and Qwen, so whether H1b holds for truthfulness depends on the operator as well as on the layer.
For sentiment in LLaMA and Qwen, the gate stays mostly closed, and from $\lambda=\tau$ to $\dmu$ the largest rotation produces a side effect below $\kappa_0$ in LLaMA and none of the directions reaches $\kappa_0$ in Qwen, so the comparison is not defined there. 

The gate also opens unequally on the two kinds of text (Table~\ref{tab:gate-open}): for $\dmu$ its mean value is about the same on neutral sentences and on task texts, but near $\lambda=\tau$ and $10\tau$ it is two to ten times larger on task texts than on neutral text in LLaMA and Qwen.
Matching $\theta$ to $\kappa_0$ on neutral text therefore allows the intermediate directions a larger intervention on the task than $\dmu$ receives at the same budget, which favors them in the H1b comparison; the rotation results support H1a, that probing and steering favor different regions, but their H1b verdicts mix the geometry of rotation with the gate and should not be read on their own.
An ungated rotation, which would separate the two, was not run.

\begin{table}[h]
\centering
\small
\begin{tabular}{llcccccc}
\hline
Concept & Model & \multicolumn{2}{c}{$\lambda=\tau$} & \multicolumn{2}{c}{$10\tau$} & \multicolumn{2}{c}{$\dmu$}\\
 & & neutral & task & neutral & task & neutral & task\\
\hline
Truthfulness & LLaMA & 0.20 & 0.33 & 0.21 & 0.43 & 0.57 & 0.64\\
 & Qwen & 0.07 & 0.19 & 0.04 & 0.16 & 0.89 & 0.67\\
 & Gemma & 0.13 & 0.15 & 0.16 & 0.18 & 1.00 & 1.00\\
Sentiment & LLaMA & 0.13 & 0.38 & 0.03 & 0.32 & 0.00 & 0.00\\
 & Gemma & 0.14 & 0.18 & 0.17 & 0.22 & 1.00 & 1.00\\
\hline
\end{tabular}
\caption{Mean value of the vMF gate $g$ on the 200 neutral sentences and on the task texts, per direction, in the rotation runs. Near $\tau$ the gate opens more on task than on neutral text; for $\dmu$ it opens equally.}
\label{tab:gate-open}
\end{table}

\section{Decision rules for RQ1}
\label{app:rules}

\paragraph{H1a.}
Let
\[
\mathcal{P}_R
=
\{\lambda:R(\lambda)\ge R(\hat\lambda_R)-\mathrm{se}_R(\lambda)\},
\]
where $\hat\lambda_R=\arg\max_\lambda R(\lambda)$ and $\mathrm{se}_R(\lambda)$ is the paired-bootstrap standard error of the difference $R(\lambda)-R(\hat\lambda_R)$, the plateau of Section~\ref{sec:rq1}; it is zero at $\lambda=\hat\lambda_R$, so the peak always belongs to the plateau.
Define $\mathcal{P}_S$ analogously.
The plateaus are sets of grid points, not necessarily contiguous; their endpoints are the smallest and largest member in grid order, and $\hat\lambda_R$ is held fixed across the bootstrap resamples when $\mathrm{se}_R$ is computed.
H1a holds if both endpoints of $\mathcal{P}_S$ are no smaller than those of $\mathcal{P}_R$, with at least one strict inequality; the rule therefore allows the two plateaus to overlap.
If $R$ is flat over the whole grid, H1a instead requires $S$ to be non-flat and $\mathcal{P}_S$ to contain $\dmu$.
The grid is capped where $\cos(w_\lambda,\dmu)>0.999$ and includes $\dmu$ itself; the cap never took effect in any run.
In the six main runs both plateaus are contiguous and do not overlap (Table~\ref{tab:matched}).
In the layer sweep of Appendix~\ref{app:layers}, the plateaus overlap in 5 of the 60 combinations; H1a holds there by the strict inequality at one endpoint in 4 of them and fails in the fifth, so 55 of the 59 cases that hold have disjoint plateaus.

\begin{figure}[h]
\centering
\includegraphics[width=\linewidth]{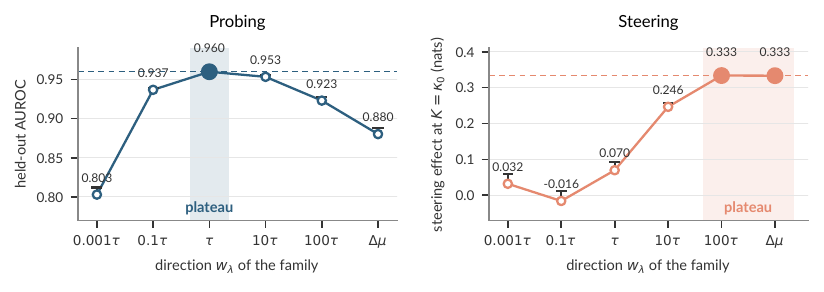}
\caption{The one-standard-error plateau rule on a real run: truthfulness in LLaMA at layer 14, from the layer sweep of Appendix~\ref{app:layers} (6-point grid). Left: held-out AUROC along the family; right: steering effect at the side effect $\kappa_0$. The dashed line marks the best value, the black tick above each point its one standard error of the gap to the best point, filled large markers the plateau members, and the shaded band the plateau. The steering plateau ($100\tau$--$\dmu$) lies to the right of the probing plateau ($\tau$), so H1a holds.}
\label{fig:plateau-rule}
\end{figure}

In plain terms, the plateau of a curve is its top together with every grid point that the data cannot tell apart from the top.
Figure~\ref{fig:plateau-rule} shows the rule on one run.
For probing, the best point is $\tau$ (AUROC 0.960); the next point, $10\tau$, lies 0.0065 below it and the standard error of that gap is 0.0023, so $10\tau$ is clearly below the top and the plateau is $\{\tau\}$.
For steering, the best point is $100\tau$ (0.333 nats); $\dmu$ lies 0.0003 below it with a standard error of 0.0063, so the two cannot be told apart and the plateau is $\{100\tau,\dmu\}$.
H1a then compares only the positions of the two plateaus on the $\lambda$ axis: the steering plateau lies to the right of the probing plateau, so H1a holds.
Two consequences of the rule explain the unusual cases in Appendix~\ref{app:layers}.
A flat curve gives a wide plateau: when all steering effects are close to zero, as for truthfulness at layer 6 of Gemma, almost every point lies within one standard error of the best one.
A plateau can also be a single point even when the neighbouring points look close, as in the probing panel, because the standard error of a paired difference is much smaller than the spread of the values themselves.

\paragraph{H1b.}
Let $\lambda^*=\arg\max_\lambda S_{(1)}(\lambda)$ on split 1.
H1b holds if
$S_{(2)}(\lambda^*)\ge1.1S_{(2)}(\dmu)$
on split 2 and the paired 95\% bootstrap interval excludes zero.
If $\lambda^*$ lies inside a flat probing plateau, we report the resulting disagreement between H1a and H1b.

\paragraph{H1c.}
We report $\rho_{\mathrm{probe}}$ only when the target probe has AUROC at least $0.6$, and $\rho_{\mathrm{steer}}$ only when the target $\dmu$ has a nonzero effect according to its 95\% bootstrap interval.
Probe-shared means $\rho_{\mathrm{probe}}\ge0.8$ and $\rho_{\mathrm{steer}}<0.5$; steering-shared reverses these conditions; both-shared means both exceed $0.8$.
Neither-shared requires both below $0.5$ and a map whose $R^2$ on held-out concept texts is at least $0.8$ times its $R^2$ on held-out neutral sentences.
All other cases are inconclusive.

\section{RQ2 protocol}
\label{app:rq2}

The 4,000 PopQA questions, the few-shot prompt (which allows ``I don't know''), and the answer judge are described in Appendix~\ref{app:data}.
Each model produces a greedy answer $a_0$ and five samples (temperature $0.7$, top-$p=0.95$, top-$k$ as in each model's generation configuration: none for LLaMA and Qwen, $64$ for Gemma; repetition penalty $1.0$).

Each answer is judged correct, abstaining, or wrong using alias matching.
Let $r(q)$ count the non-abstaining samples that repeat $a_0$.
We use ``knowledge-available'' operationally: at least one of the six answers is correct. This shows that the model can reach the answer under some decoding, not that it explicitly knows the fact; conversely, ``no correct answer'' means only that none of the six tested answers was correct.

The probe measures whether a correct answer is reachable, not whether the model recognizes its own answer as false.
The cases follow the error taxonomy of \citet{orgad2025llms}, who resample 30 answers per question and separate, among others, consistently incorrect answers, answers spread over many alternatives, and errors among whose resampled answers the correct one also appears; our C3, C2 and C1 roughly correspond to these three types, defined here from six answers with abstention allowed.
Six answers are far fewer than their thirty, and the labels inherit a sampling bias: a question whose correct answer has probability $p$ under sampling is missed by all five samples with probability $(1-p)^5$, 0.77 at $p=0.05$ and 0.44 at $p=0.15$, and is then counted among the questions without a correct answer rather than in C1.
C1 is therefore a conservative set, and C2 contains some questions that more samples would have moved to C1; C0 is defined by the greedy answer alone and is unaffected.
This is why C1 is small (71, 33 and 35 questions) and its intervals wide, and why the 10\% limit of H2b(ii) rests on 7, 3 and 3 questions; the C2-versus-C3 contrasts, with 180--576 questions per class, are not affected in the same way. 

They probe the generated answer tokens to predict the error type and to select a correct answer among the samples.
We instead probe before any answer is generated, compare the probe with an output-only baseline, and ask what steering does to each type of error.
In our data, the entropy of the first answer token separates C2 from C3 better than the probe (Figure~\ref{fig:app-rq2-read}).

The two parts of H2a ask whether the representation tracks whether the answer is reachable at all and, when it is not, whether it tracks the model's commitment to its wrong answer.
H2b(ii) expects few corrections because steering toward ``unknown'' should produce abstention, not retrieval, and H2b(iii) tests the split between probing and steering on ``unknown'' itself.

All changes are measured against an unsteered rerun, which agrees with the labelling run on 99.3--100\% of the questions.
The budget $\kappa_0$ is 0.0012, 0.0018 and 0.0018 nats for LLaMA, Qwen and Gemma.
We use $\dmu$ at its anchor strength ($K=\kappa_0$), and the probe end, the random direction, and the transferred $\dmu$ interpolated per question to $K=\kappa_0$, linearly in $\sqrt K$ between the two strengths that bracket it, as in RQ1 (Appendix~\ref{app:measures}). Interpolating linearly in $\log K$ instead changes no estimate by more than 1\,pp and no verdict.
The gated direction is compared at the strength of $\dmu$, because its $K$ on neutral text reflects how often the probe flags neutral sentences, not its effect on questions.

\begin{table}[h]
\centering
\small
\begin{tabular}{ll}
\hline
Case & Definition ($a_0$ = greedy answer; knowledge-available = some of the six answers is correct)\\
\hline
C0 & $a_0$ correct\\
C1 & $a_0$ wrong, some answer correct: knowledge-available error\\
C2 & $a_0$ wrong, no correct answer, $r(q)\le2$: uncertain error\\
C3 & $a_0$ wrong, no correct answer, $r(q)\ge4$: confident error\\
Cm & $a_0$ wrong, no correct answer, $r(q)=3$\\
AU, AK & $a_0$ abstains; no correct answer (AU) / knowledge-available (AK)\\
INV & $a_0$ produced no answer (Qwen only; Appendix~\ref{app:data})\\
\hline
\end{tabular}
\caption{Cases used in RQ2, defined from the model's own answers.}
\label{tab:cases}
\end{table}

Probes for questions without versus with a correct answer among the six are extracted at the last prompt token and cross-fitted over two folds.
A question is flagged if its probe score exceeds the 90th percentile of C0 scores held out within the training fold.
For C2 vs.\ C3, H2a requires the bootstrap 95\% interval of the C2--C3 flag-rate difference to lie above zero.
For each contrast of H2a (C1 vs.\ $\text{C2}\cup\text{Cm}\cup\text{C3}$; C2 vs.\ C3), the probe counts as adding information beyond the output distribution only if its AUROC exceeds that of $H_{\mathrm{ent}}$, or a cross-fitted logistic combination of the two exceeds $H_{\mathrm{ent}}$ alone, with a paired bootstrap interval above zero; otherwise, we conclude that the hidden state adds no evidence beyond the output uncertainty.
Because C1 can be small when abstention is allowed, the first contrast is also reported with abstentions included ($\text{C1}\cup\text{AK}$ vs.\ $\text{C2}\cup\text{Cm}\cup\text{C3}\cup\text{AU}$).
We report over-abstention on C0 and the split of initially wrong answers into abstentions and corrections.
Across models, we also report Cohen's $\kappa$ of the knowledge-available labels, whether a transferred probe flags C2 more often than C3, and whether a transferred $\dmu$ increases abstention in the target model.

Table~\ref{tab:rq2-xmodel} carries the probe and the mean difference $\dmu$ of a source model $A$ into a target $B$ with the maps of Section~\ref{sec:transfer}.
Its column $\kappa$ is Cohen's $\kappa$ between the two models' labels ``some correct answer''.
C3 overlap is the share of $A$'s confident errors that are also confident errors of $B$; it is not symmetric because the models have different numbers of C3 questions (259 in LLaMA, 83 in Qwen, 576 in Gemma).
Probe AUROC is the out-of-fold AUROC in $B$, for ``no correct answer'' versus ``some correct answer'', of $A$'s probe carried by the probe law, with $B$'s own probe in brackets, and $\rho_{\mathrm{probe}}$ is the share of the own probe's above-chance AUROC that the carried probe keeps (Eq.~\eqref{eq:rho}).

C2$-$C3 flags is the share of $B$'s C2 questions minus the share of its C3 questions that the probe flags, in percentage points, for the carried probe and, in brackets, for $B$'s own probe; these are point estimates without intervals.

The column $\cos$ is the cosine between $A$'s $\dmu$ carried by the steering law and $B$'s own $\dmu$.
C2 gain is the rise in abstention on $B$'s uncertain errors, in percentage points, for the carried $\dmu$ and, in brackets, for $B$'s own $\dmu$, both at $B$'s budget $\kappa_0$, and $\rho_{\mathrm{steer}}$ is their ratio.
The intervals are 95\% bootstrap intervals over questions (2,000 resamples; for $\rho_{\mathrm{steer}}$ over $B$'s C2 questions).

\begin{table}[h]
\centering
\footnotesize
\setlength{\tabcolsep}{2.5pt}
\resizebox{\textwidth}{!}{%
\begin{tabular}{@{}lcccccccc@{}}
\hline
 & & C3 overlap & Probe AUROC & C2$-$C3 flags, pp & & & C2 gain, pp & \\
$A\rightarrow B$ & $\kappa$ & (\%) & transf.\ (own) & transf.\ (own) & $\rho_{\mathrm{probe}}$ [95\% CI] & $\cos$ & transf.\ (own) & $\rho_{\mathrm{steer}}$ [95\% CI]\\
\hline
L $\rightarrow$ Q & 0.56 & 17 & 0.910 (0.945) & 15 (32) & 0.92 [0.90, 0.94] & $+0.35$ & 12.0 (30.1) & 0.40 [0.26, 0.56]\\
L $\rightarrow$ G & 0.60 & 51 & 0.897 (0.933) & 25 (21) & 0.92 [0.90, 0.94] & $+0.56$ & 10.4 (18.3) & 0.57 [0.45, 0.71]\\
Q $\rightarrow$ L & 0.56 & 53 & 0.870 (0.915) & 27 (28) & 0.89 [0.87, 0.91] & $+0.24$ & 13.2 (44.4) & 0.30 [0.21, 0.40]\\
Q $\rightarrow$ G & 0.65 & 63 & 0.892 (0.933) & 29 (21) & 0.91 [0.89, 0.92] & $-0.36$ & 7.3 (18.3) & 0.40 [0.27, 0.54]\\
G $\rightarrow$ L & 0.60 & 23 & 0.893 (0.915) & 32 (28) & 0.95 [0.93, 0.96] & $+0.29$ & 14.4 (44.4) & 0.32 [0.23, 0.42]\\
G $\rightarrow$ Q & 0.65 & 9 & 0.928 (0.945) & 18 (32) & 0.96 [0.95, 0.98] & $+0.54$ & 23.9 (30.1) & 0.79 [0.66, 0.92]\\
\hline
\end{tabular}}
\caption{Agreement across models on ``unknown'' for each source $A$ and target $B$ (L = LLaMA, Q = Qwen, G = Gemma), without a hypothesis fixed in advance. Probes keep 89--96\% of their performance and flag C2 15--32\,pp more often than C3; $\dmu$ keeps 30--79\% of its C2 gain.}
\label{tab:rq2-xmodel}
\end{table}

For ``unknown'', the probe chosen by cross-validation lies at $3\tau$ in at least one fold of every model and at up to $32\tau$ in the other, near the best sentiment probe ($3\tau$ in LLaMA and Qwen), and it steers 73--100\% as strongly as $\dmu$ on uncertain errors.
In Table~\ref{tab:rq2-xmodel}, 9 to 63\% of the confident errors (C3) of one model are also confident errors of another.
As for sentiment in RQ1, the mean difference carried from Qwen into Gemma points away from Gemma's own ($\cos=-0.36$; $-0.40$ and $-0.32$ in the two folds) and still raises abstention on its uncertain errors by 7\,pp, 40\% of the gain of Gemma's own $\dmu$.
It points away from Gemma's raw $\dmu$ but not from Gemma's probe: its cosine with $w_{\lambda^*}$ is $+0.11$ and $+0.12$, and with the random direction $0.01$ and $-0.02$.
A sign-flipped carried vector was not run; the random direction at the same $K$, which raises abstention on Gemma's uncertain errors by 1\,pp against 7\,pp for the carried $\dmu$ (Section~\ref{sec:res-rq2}), bounds the part of that gain that any push would produce.

\section{Sensitivity of the gate threshold}
\label{app:gate}

The gate of H2c flags a question when its probe score exceeds the 90th percentile of the C0 scores on an inner split of the training fold.
To see how the verdicts depend on this choice, we recompute post hoc the gated outcomes for thresholds between the 75th and the 97.5th percentile of the C0 scores on the whole training fold; the original threshold lies at its 91st to 96th percentile.
No new answers are generated.
The gated run equals, question by question, the $\dmu$-steered answer where a question is flagged and the unsteered answer elsewhere (agreement 99.7--100\%), so another threshold only changes which of the two saved answers is used.

\begin{table}[h]
\centering
\small
\begin{tabular}{lccc}
\hline
Threshold (C0 percentile) & LLaMA & Qwen & Gemma\\
\hline
original (inner split, 90th) & 0.65 / 0.33 \,fails & 0.84 / 0.29 \,holds & 0.95 / 0.80 \,fails\\
75th & 0.91 / 0.63 \,fails & 0.95 / 0.64 \,fails & 0.98 / 1.00 \,fails\\
80th & 0.86 / 0.59 \,fails & 0.97 / 0.57 \,fails & 0.98 / 1.00 \,fails\\
85th & 0.79 / 0.54 \,fails & 0.92 / 0.43 \,holds & 0.98 / 0.80 \,fails\\
90th & 0.70 / 0.41 \,fails & 0.86 / 0.36 \,holds & 0.98 / 0.80 \,fails\\
95th & 0.61 / 0.22 \,fails & 0.81 / 0.36 \,holds & 0.92 / 0.80 \,fails\\
97.5th & 0.54 / 0.19 \,fails & 0.59 / 0.29 \,fails & 0.87 / 0.60 \,fails\\
\hline
\end{tabular}
\caption{Gating at the strength of $\dmu$ for different thresholds: share of $\dmu$'s abstention gain on C2 kept / share of its cost on C0 paid, and the H2c verdict (kept $\ge0.8$ and paid $\le0.5$).}
\label{tab:gate}
\end{table}

Table~\ref{tab:gate} shows the trade-off.
In LLaMA, H2c fails at every threshold because a threshold loose enough to keep 80\% of the gain also pays more than half of the cost.
In Qwen, it holds from the 85th to the 95th percentile and fails at looser and stricter thresholds.
In Gemma, it fails at every threshold, but there the cost of $\dmu$ on C0 is only +0.8\,pp, about five of its 644 C0 questions, so the share paid rests on one or two questions.

\begin{table}[h]
\centering
\footnotesize
\setlength{\tabcolsep}{4pt}
\begin{tabular}{llcccc}
\hline
 & & ungated $\dmu$ & probe-gated & entropy-gated & ungated $\dmu$ at the\\
Model & & & & & probe gate's C0 cost\\
\hline
LLaMA & C0, pp (questions) & $+7.5$ (63) & $+2.5$ (21) & $+3.4$ (29) & $+2.5$\\
 & C2, pp & $+44.4$ & $+28.9$ & $+40.0$ & $+24.9$\\
 & kept / paid & -- & 0.65 / 0.33 & 0.90 / 0.46 & --\\
Qwen & C0, pp (questions) & $+3.0$ (14) & $+0.8$ (4) & $+1.3$ (6) & $+0.8$\\
 & C2, pp & $+30.1$ & $+25.2$ & $+22.8$ & $+12.2$\\
 & kept / paid & -- & 0.84 / 0.29 & 0.76 / 0.43 & --\\
Gemma & C0, pp (questions) & $+0.8$ (5) & $+0.6$ (4) & $+0.6$ (4) & $+0.6$\\
 & C2, pp & $+18.3$ & $+17.4$ & $+18.0$ & $+15.0$\\
 & kept / paid & -- & 0.95 / 0.80 & 0.98 / 0.80 & --\\
\hline
\end{tabular}
\caption{H2c with two gates and at matched cost, all with $\dmu$ at its anchor strength, composed from the saved runs. C0: over-abstention on correct answers, in percentage points, with the number of questions in brackets (845, 472 and 644 C0 questions); C2: abstention gained on uncertain errors; kept and paid as in the H2c rule (gain kept $\ge0.8$, cost paid $\le0.5$). The last column is ungated $\dmu$ at the strength whose C0 cost equals the probe gate's.}
\label{tab:gate-baselines}
\end{table}

The entropy gate (Table~\ref{tab:gate-baselines}) flags a question when its first-token entropy exceeds the 90th percentile of the C0 entropies of the training fold, cross-fitted as for the probe; it flags 78, 63 and 88\% of C2 and 22, 24 and 41\% of C3.
The matched-cost comparison reads the ungated $\dmu$ curve, C2 gain against C0 cost over the four strengths, at the C0 cost of the gated run by linear interpolation; in Qwen the gated outcomes are composed from runs that agree with the saved gated run on 99.65\% of the questions, about 14 questions, which is of the same order as the C0 counts there.

\section{Datasets and prompts}
\label{app:data}

All datasets are public (Table~\ref{tab:data}): TruthfulQA \citep{lin2022truthfulqa}, Yelp polarity \citep{zhang2015character-yelp-polarity}, WikiText-2 \citep{merity2017pointer}, and PopQA \citep{mallen2023popqa}. Every model receives the same texts, so examples line up across models although the tokenizers differ; activations are taken at the last token of each text. All splits and samples use seed 0.

\begin{table}[h]
\centering
\small
\begin{tabular}{>{\raggedright\arraybackslash}p{1.8cm}>{\raggedright\arraybackslash}p{2.9cm}>{\raggedright\arraybackslash}p{5.0cm}>{\raggedright\arraybackslash}p{2.3cm}}
\hline
Dataset & Used for & Source and selection & Size used\\
\hline
TruthfulQA & truthfulness: probing (MC2) and steering (MC1) & all questions; for probing, each question paired with each of its MC2 answers; for steering, its MC1 answer list & 817 questions; 5,882 MC2 and 4,114 MC1 answers\\
Yelp polarity & sentiment: probing and steering & test split; random sample per class; the probe reads the first 120 words, the steering prompt the first 24 &  1,000 positive, 1,000 negative\\
WikiText-2 & side effect $K$ & raw test split; first 200 sentences with at least 15 words & 200 sentences\\
WikiText-2 & cross-model maps & raw test split; random sentences with at least 6 words, outside the side-effect set & 8,000 sentences, each as its first half and in full (16,000 texts: 12,800 for fitting, 3,200 held out)\\
PopQA & RQ2 & test split; ten relations, $s_{\mathrm{pop}}>0$, 1,000 questions per popularity quartile & 4,000 questions\\
\hline
\end{tabular}
\caption{Datasets, their role, and the amount used.}
\label{tab:data}
\end{table}

\paragraph{TruthfulQA.}
TruthfulQA consists of 817 questions; the multiple-choice lists of its release contain 5,882 reference answers (2,585 true, 3,297 false).
The dataset release provides two multiple-choice lists per question.
The multi-true list (MC2) contains all reference answers of the question, of which one or more are true; it corresponds to the multiple-choice task of \citet[Sec.~3.2]{lin2022truthfulqa}, while the names MC1 and MC2 and the answer counts come from the dataset release.
The single-true list (MC1) contains the same false answers but only one true answer (4,114 answers, 2--13 per question).
For MC1, a model selects the answer to which it assigns the highest likelihood, and the score is the share of questions for which this answer is the true one.
For MC2, the score is the total probability the model assigns to the true answers, normalized over all answers in the list.
We use the MC2 answers as labeled probing examples and the MC1 list for the steering margin; we do not report the MC1 or MC2 scores themselves.
A probing example is the text \verb|Q: {question} A: {answer}| for one MC2 answer, labeled 1 if the answer is true and 0 otherwise.
The split is by question, so no question contributes to both estimation and evaluation: 408 questions (2,944 answers) form the estimation subset, and 204 and 205 questions (2,938 answers in total) form the two held-out sets.
Steering is measured on the 409 held-out questions with their MC1 answers.
The prompt is the ``helpful'' instruction of \citet[App.~E]{lin2022truthfulqa}, which asks the model to interpret each question literally and to reply ``I have no comment'' unless it is completely certain, followed by \verb|Q: {question} A: {answer}|.
The MC1 margin is the log-likelihood of the correct answer minus the largest log-likelihood among the incorrect answers, each summed over the answer tokens only; $S$ is its change under steering.

\paragraph{Yelp polarity.}
We draw 1,000 positive and 1,000 negative reviews from the test split and split them 50/25/25 by review.
A probing example is the first 120 words of a review.
A steering prompt is the first 24 words of a held-out review followed by ``Overall, the experience was''.
The steering effect is the change in $\log\sum_{v\in P}e^{z_v}-\log\sum_{v\in Q}e^{z_v}$ for the next-token logits $z$, where $P$ and $Q$ are the first tokens of eight positive words (great, good, excellent, amazing, wonderful, fantastic, perfect, pleasant) and eight negative words (terrible, bad, awful, horrible, poor, disappointing, mediocre, unpleasant).

\paragraph{WikiText-2.}
The side-effect set $N$ consists of the first 200 sentences of the raw test split (9,360 sentences) with at least 15 words; $K$ is computed from the next-token distribution at the last token of each sentence.
The maps use 8,000 further sentences with at least 6 words, drawn at random from the 9,065 other sentences with at least 6 words.
Each sentence gives two texts, its first half (cut at a word boundary) and the full sentence, so that the last token falls on the same word in every model.
A random 20\% of these sentences is held out; the ridge strength is chosen by the $R^2$ on them, and this $R^2$ is the neutral reference in the neither-shared rule (Appendix~\ref{app:rules}).
No concept text is used to fit a map.

\paragraph{PopQA.}
From the test split of PopQA \citep{mallen2023popqa} we keep ten relations and questions with subject popularity $s_{\mathrm{pop}}>0$, and draw 1,000 questions from each quartile of $\log_{10}s_{\mathrm{pop}}$ (boundaries 2.49, 3.11 and 3.87), out of 10,359 questions that meet these conditions.
The sample contains director (774 questions), screenwriter (773), producer (593), author (590), composer (374), capital (244), place of birth (229), father (219), capital of (131), and mother (73). (``capital'' asks for the capital of a region, ``capital of'' for the region a city is the capital of; both are PopQA's own relation labels).
The prompt, without a chat template, is shown below; the third example is a fictional person, so that ``I don't know'' is demonstrated as an answer.
{\scriptsize
\begin{verbatim}
Answer each question with a short answer. If you do not know the answer, reply "I don't know".
Q: Who directed the film Jaws?
A: Steven Spielberg
Q: What is the capital of Australia?
A: Canberra
Q: Who is the father of Zelvorin Takamashi?
A: I don't know
Q: {question}
A:
\end{verbatim}
}

\noindent Answers are generated with at most 24 new tokens, and the answer is the first non-empty line of the continuation, cut before any new \texttt{Q:}.
After normalization (lowercase, punctuation and articles removed), an answer abstains if it is empty or contains a phrase from a fixed list (for example ``I don't know'', ``not sure'', ``no idea''); otherwise it is correct if it contains one of PopQA's possible answers (answers of at most three characters must match a whole word), and wrong otherwise.

Qwen3.5 sometimes opens a reasoning block (\texttt{<think>}) instead of answering. For such continuations, we generate up to 384 new tokens and take the answer from the text after the closing \texttt{</think>}; if the block does not close within this budget, or the answer is only a chat role word, the answer is invalid, which counts as neither correct, wrong, nor an abstention.

This leaves 121 of Qwen's 4,000 greedy answers (3\%) and 187 of its 20,000 samples invalid; LLaMA and Gemma produce no invalid answers. Questions whose greedy answer is invalid form their own case (INV) and are not part of C0--C3.
For the entropy baseline, Qwen's first answer token is taken after a closed reasoning block (107 questions); for 123 questions whose block does not close, the entropy stays at the first generated token.

The steered runs use the same budget, so a steered continuation that opens a reasoning block is also extended; invalid continuations count as neither an abstention nor a correct answer, and at the anchor strength $\dmu$ leaves 108 of them (17 among the 123 C2 questions) against 124 in the unsteered rerun.
Reading $H_{\mathrm{ent}}$ after a closed block puts it at a different position than the probe for 107 questions; excluding them, its AUROC on C2 against C3 rises from 0.69 to 0.84 and on the H2a(i) contrasts from 0.59 to 0.62 and from 0.73 to 0.74, so this does not favor the probe in any contrast.
\section{Reading position: last token versus pooling}
\label{app:pool}

A probe is obtained in two steps.
First, estimation: from the hidden states of the training texts, $\dmu$ and $\SigW$ are computed and from them the direction $w$ (Sections~\ref{sec:directions} and~\ref{sec:eval}); this is done once, and the result is a fixed vector.
Second, scoring: for a held-out text, some of its hidden states are multiplied by $w$ and the products are combined into one number, the probe score, from which the AUROC is computed.
In all our main experiments both steps use the hidden state at the last token of the text.
AxBench \citep[Sec.~3.2]{wu2025axbench} instead scores every token of a text and keeps the maximum.
We compare three ways of turning a text into one probe score.

The first is the one used in all our main experiments: the direction $w$ is estimated from the hidden states at the last token of the texts, and a held-out text is scored by $w^\top h_T$ at its last token.
The second is max pooling, the scoring used by AxBench \citep[Sec.~3.2]{wu2025axbench}: every token of the text is scored, $w^\top h_t$ for $t=1,\dots,T$, and the largest score is kept;

In the second variant, the vector $w$ is computed once from last-token states, as in the first variant (ours), so the direction is the same in both variants.
It is then, however, multiplied with the hidden state of every token and the largest product is kept, so the same text can receive a different score in the two variants.
The hidden states at the intermediate tokens have their own class means and within-class spread, and a direction estimated from them would differ from the one estimated at the last token; no variant estimates $\dmu$ and $\SigW$ from all token positions. 

The third is mean pooling: the hidden states of the text are averaged into $\bar h$, $\dmu$ and $\SigW$ are re-estimated from these averaged states, and a held-out text is scored by $w^\top\bar h$; here both the estimation and the scoring change.
In the second variant the direction estimated at the last token is applied to every token, whose class means and within-class covariance can differ from those at the last token; no variant estimates $\dmu$ and $\SigW$ from all token positions.

Only the tokens that carry the concept are pooled: the answer for TruthfulQA and the whole review for sentiment; special tokens are excluded.
Each variant chooses its best direction of the family on the first held-out split, and the choices are compared on the second split with a paired bootstrap that resamples questions or reviews (1,000 resamples).
Significance is corrected with Holm's method over all 18 comparisons we ran, including a third variant, directions computed from mean states but scored by max pooling, which is omitted here and was never better than the last token.

Table~\ref{tab:pool} gives the held-out AUROC of the last-token reading on the second held-out split and, for each pooling, the difference to it (negative: the last token is better), with paired bootstrap 95\% intervals.
After correction, the last token was significantly better than the alternative in 8 of the 12 comparisons and never significantly worse. Max pooling, the scoring used by AxBench, was worse in all six cases and lost up to 0.10 AUROC on truthfulness.
Mean pooling gave a slightly higher AUROC for sentiment in all three models ($+0.006$ to $+0.015$), but none of these differences are significant after correction.
Sentiment is spread over the whole review rather than concentrated at its end, which may be why averaging does not hurt there, while it lowers the AUROC for truthfulness. The loss under max pooling therefore shows that a direction fitted at the last token does not transfer to scoring every token, not that token-level estimation would do worse; directions estimated across all pooled positions were not run.

\begin{table}[h]
\centering
\small
\setlength{\tabcolsep}{4pt}
\begin{tabular}{llccc}
\hline
Concept & Model & Last token & Max pooling $-$ last token & Mean pooling $-$ last token\\
\hline
Truthfulness & LLaMA & 0.957 & $-0.051$ [$-0.066$, $-0.036$]$^{*}$ & $-0.016$ [$-0.026$, $-0.007$]$^{*}$\\
 & Qwen & 0.957 & $-0.092$ [$-0.114$, $-0.073$]$^{*}$ & $-0.007$ [$-0.015$, $+0.001$]\\
 & Gemma & 0.969 & $-0.101$ [$-0.126$, $-0.078$]$^{*}$ & $-0.013$ [$-0.023$, $-0.005$]$^{*}$\\
\hline
Sentiment & LLaMA & 0.983 & $-0.043$ [$-0.063$, $-0.025$]$^{*}$ & $+0.006$ [$-0.001$, $+0.015$]\\
 & Qwen & 0.968 & $-0.078$ [$-0.106$, $-0.050$]$^{*}$ & $+0.015$ [$+0.003$, $+0.029$]\\
 & Gemma & 0.982 & $-0.054$ [$-0.076$, $-0.034$]$^{*}$ & $+0.009$ [$+0.001$, $+0.019$]\\
\hline
\end{tabular}
\caption{Held-out AUROC of the last-token reading and its difference to max and mean pooling, with 95\% intervals ($^{*}$: significant after Holm correction). Max pooling is worse in every case.}
\label{tab:pool}
\end{table}

\section{Layer choice basis}
\label{app:layers}

The paper reads and steers every model at about $0.44$ of its depth (Section~\ref{sec:models-data}).
This appendix gives the basis for that choice: how well each concept can be probed, and how strongly $\dmu$ steers it, at other depths.
To check this, we repeat the RQ1 measurements at ten relative depths, from $1/8$ to $7/8$ of the layers: layers 4, 8, 10, 12, 14, 16, 18, 20, 24 and 28 of 32 in LLaMA and Qwen, and 6, 12, 15, 18, 21, 24, 27, 30, 36 and 42 of 48 in Gemma.
The data, splits, and steering prompts are those of Section~\ref{sec:eval}, and the probe and the steering intervention always use the same layer.

To keep the computing time of the sweep manageable, with ten depths for each model and concept and six steering strengths per direction, the family has six members instead of fourteen, $\lambda\in\{10^{-3},0.1,1,10,100\}\,\tau$ and $\dmu$.
Values at the paper's layer can therefore differ slightly from those of Section~\ref{sec:res-rq1} when the main run's best direction lies at a grid point the sweep lacks: the best AUROC does so only for sentiment in LLaMA (0.980 against 0.981) and Qwen (0.964 against 0.966), whose best probe sits at $3\tau$, and the H1b ratios, whose $\lambda^*$ is chosen on the coarser grid, by up to 0.08 (truthfulness in Gemma, 0.99 against 1.07).

Each layer has its own budget $\kappa_0$, the $K$ of $\dmu$ at $a=0.1$ ($0.01$ for Gemma) at that layer.
For RQ2, we sweep probing only.
The PopQA labels do not depend on the layer, so we read the prompts of all 4,000 questions at the ten layers and compute the cross-fitted probe of Section~\ref{sec:eval-know} at each of them, with 95\% bootstrap intervals (200 resamples).
Steering toward ``unknown'' was not swept.

\begin{table}[h]
\centering
\small
\setlength{\tabcolsep}{3pt}
\begin{tabular}{llccccccccc}
\hline
 & & \multicolumn{3}{c}{Truthfulness} & \multicolumn{3}{c}{Sentiment} & \multicolumn{3}{c}{PopQA (``unknown'')}\\
Depth & Layer & LLaMA & Qwen & Gemma & LLaMA & Qwen & Gemma & LLaMA & Qwen & Gemma\\
\hline
0.125 & 4 / 6 & 0.886 & 0.874 & 0.821 & 0.915 & 0.880 & 0.831 & 0.782 & 0.764 & 0.650\\
0.25 & 8 / 12 & 0.931 & 0.916 & 0.894 & 0.957 & 0.920 & 0.922 & 0.888 & 0.918 & 0.885\\
0.3125 & 10 / 15 & 0.946 & 0.932 & 0.927 & 0.971 & 0.943 & 0.957 & 0.902 & 0.935 & 0.859\\
0.375 & 12 / 18 & 0.958 & 0.938 & 0.947 & 0.979 & 0.960 & 0.975 & 0.911 & 0.942 & 0.929\\
0.4375$^{\dagger}$ & 14 / 21 & \textbf{0.960} & 0.961 & 0.968 & \textbf{0.980} & 0.964 & \textbf{0.978} & 0.915 & 0.945 & 0.933\\
0.5 & 16 / 24 & 0.959 & \textbf{0.966} & \textbf{0.971} & 0.977 & \textbf{0.967} & 0.977 & 0.918 & 0.947 & \textbf{0.937}\\
0.5625 & 18 / 27 & 0.952 & 0.965 & 0.964 & 0.974 & \textbf{0.967} & 0.977 & 0.919 & \textbf{0.948} & 0.932\\
0.625 & 20 / 30 & 0.948 & 0.961 & 0.959 & 0.970 & 0.964 & 0.972 & 0.919 & 0.948 & 0.928\\
0.75 & 24 / 36 & 0.939 & 0.955 & 0.953 & 0.963 & 0.957 & 0.969 & 0.917 & 0.941 & 0.925\\
0.875 & 28 / 42 & 0.937 & 0.945 & 0.949 & 0.964 & 0.955 & 0.967 & \textbf{0.921} & 0.943 & 0.915\\
\hline
\end{tabular}
\caption{Held-out AUROC of the best direction of the family at each relative depth ($^{\dagger}$: the depth used in the paper; bold: the best depth). From $3/8$ of the depth upward, probing varies little.}
\label{tab:sweep-probe}
\end{table}

Table~\ref{tab:sweep-probe} gives, for each relative depth, the layer in LLaMA and Qwen and in Gemma and the held-out AUROC of the best direction of the family, on all 4,000 questions for PopQA.
From $3/8$ of the depth upward, the AUROC of each model and concept varies by at most 0.04 between layers.
Averaged over truthfulness, sentiment, and PopQA, the best layer is the paper's layer 14 in LLaMA, layer 18 in Qwen, and layer 24 in Gemma; the layers we use come within 0.004 AUROC of that average in Qwen (0.957 against 0.960) and within 0.002 in Gemma (0.960 against 0.962).
On PopQA, the AUROC stays within 0.01 of its maximum from $3/8$ of the depth upward in LLaMA and Qwen, and within 0.03 in Gemma, whose layer 21 lies 0.004 below its best layer, 24.

In all three models and for all three concepts, the 95\% interval of the AUROC at the paper's layer overlaps that of the best layer (Table~\ref{tab:sweep-ci}); the largest gap is 0.006 AUROC (PopQA in LLaMA).
For truthfulness and sentiment, the sweep stored no per-example scores, so these intervals use the normal approximation of \citet{hanley1982auc} (1,291 true and 1,647 false held-out answers; 500 positive and 500 negative held-out reviews).

Two caveats apply to these intervals.
They treat the held-out answers as independent, but the TruthfulQA answers come in groups of about seven per question, so for truthfulness they are narrower than a question-clustered bootstrap would give; they are descriptive and enter no hypothesis test.
The best direction of each layer is also chosen on the same held-out data it is reported on, which favors every layer alike.
Choosing it on the first held-out split and reporting it on the second instead changes nothing of substance: the AUROC at the paper's layer is then within 0.006 of the best layer's in every model and concept (0.961 against 0.967 at layer 16 for truthfulness in Qwen, 0.969 against 0.974 at layer 24 in Gemma, at most 0.004 elsewhere), and the best layer is unchanged except for sentiment in LLaMA, where layer 16 exceeds layer 14 by 0.002.

\begin{table}[h]
\centering
\small
\begin{tabular}{llcc}
\hline
Concept & Model & Paper's layer: AUROC [95\% CI] & Best layer: AUROC [95\% CI]\\
\hline
Truthfulness & LLaMA & 14: 0.960 [0.952, 0.968] & same layer\\
 & Qwen & 14: 0.961 [0.954, 0.969] & 16: 0.966 [0.959, 0.973]\\
 & Gemma & 21: 0.968 [0.961, 0.975] & 24: 0.971 [0.964, 0.978]\\
\hline
Sentiment & LLaMA & 14: 0.980 [0.971, 0.989] & same layer\\
 & Qwen & 14: 0.964 [0.952, 0.976] & 18: 0.967 [0.956, 0.979]\\
 & Gemma & 21: 0.978 [0.969, 0.987] & same layer\\
\hline
PopQA (``unknown'') & LLaMA & 14: 0.915 [0.906, 0.923] & 28: 0.921 [0.913, 0.929]\\
 & Qwen & 14: 0.945 [0.936, 0.953] & 18: 0.948 [0.938, 0.956]\\
 & Gemma & 21: 0.933 [0.924, 0.945] & 24: 0.937 [0.930, 0.948]\\
\hline
\end{tabular}
\caption{Held-out AUROC of the best direction at the paper's layer and at the best layer of the sweep, with 95\% intervals. The intervals overlap in every case.}
\label{tab:sweep-ci}
\end{table}

\begin{table}[h]
\centering
\small
\setlength{\tabcolsep}{2.5pt}
\begin{tabular}{llcccccccccccc}
\hline
 & & \multicolumn{6}{c}{Steering effect of $\dmu$ at $K=\kappa_0$} & \multicolumn{6}{c}{H1b: $S_{(2)}(\lambda^*)/S_{(2)}(\dmu)$}\\
 & & \multicolumn{3}{c}{Truthfulness} & \multicolumn{3}{c}{Sentiment} & \multicolumn{3}{c}{Truthfulness} & \multicolumn{3}{c}{Sentiment}\\
Depth & Layer & L & Q & G & L & Q & G & L & Q & G & L & Q & G\\
\hline
0.125 & 4 / 6 & -0.016 & 0.063 & 0.014 & 0.071 & 0.040 & 0.002 & -- & 1.78 & 0.45 & 1.62$^{*}$ & 1.97$^{*}$ & --\\
0.25 & 8 / 12 & 0.309 & 0.196 & 0.125 & 0.228 & 0.229 & 0.086 & 0.97 & 1.12 & 1.28 & 1.42$^{*}$ & 1.68$^{*}$ & 1.70$^{*}$\\
0.3125 & 10 / 15 & 0.455 & 0.269 & 0.246 & 0.300 & 0.364 & 0.253 & 1.00 & 1.18$^{*}$ & 1.29$^{*}$ & 1.46$^{*}$ & 1.33$^{*}$ & 1.36$^{*}$\\
0.375 & 12 / 18 & 0.427 & 0.308 & 0.368 & 0.457 & 0.460 & 0.515 & 1.03 & 1.27$^{*}$ & 0.85 & 1.41$^{*}$ & 1.56$^{*}$ & 1.26$^{*}$\\
0.4375$^{\dagger}$ & 14 / 21 & 0.333 & 0.600 & 0.561 & 0.849 & 0.589 & 0.732 & 0.99 & 1.10 & 0.99 & 1.29$^{*}$ & 1.50$^{*}$ & 1.27$^{*}$\\
0.5 & 16 / 24 & 0.194 & 0.562 & 0.381 & 1.109 & 0.668 & 0.967 & 1.27$^{*}$ & 1.05 & 1.11 & 1.29$^{*}$ & 1.53$^{*}$ & 1.33$^{*}$\\
0.5625 & 18 / 27 & 0.117 & 0.289 & 0.046 & 1.065 & 1.050 & 0.931 & 1.57$^{*}$ & 1.43$^{*}$ & 0.44 & 1.28$^{*}$ & 1.37$^{*}$ & 1.39$^{*}$\\
0.625 & 20 / 30 & 0.065 & 0.121 & 0.047 & 1.097 & 1.041 & 0.772 & 1.72 & 2.41$^{*}$ & 0.44 & 1.29$^{*}$ & 1.29$^{*}$ & 1.42$^{*}$\\
0.75 & 24 / 36 & 0.041 & 0.096 & 0.030 & 0.974 & 1.266 & 0.627 & 1.52 & 2.32$^{*}$ & -0.80 & 1.31$^{*}$ & 1.46$^{*}$ & 1.34$^{*}$\\
0.875 & 28 / 42 & 0.011 & 0.136 & 0.033 & 1.135 & 1.359 & 0.564 & 1.92 & 1.32$^{*}$ & -0.94 & 1.23$^{*}$ & 1.42$^{*}$ & 1.43$^{*}$\\
\hline
\end{tabular}
\caption{Steering effect of $\dmu$ at each layer's $\kappa_0$ (left) and the H1b ratio (right) across depths (L = LLaMA, Q = Qwen, G = Gemma; $^{*}$: H1b holds; $^{\dagger}$: the paper's depth). H1b holds for sentiment wherever $\dmu$ steers, for truthfulness only at some layers.}
\label{tab:sweep-steer}
\end{table}

The left half of Table~\ref{tab:sweep-steer} gives the steering effect of $\dmu$ at each layer's own budget $\kappa_0$, in the units of each concept (change of the answer margin in nats; change of the log-odds of positive versus negative words).

The right half gives the effect of the direction $\lambda^*$ chosen on the first held-out split, relative to $\dmu$, on the second split; H1b holds when this ratio is at least 1.1 and the paired 95\% interval of the difference excludes zero.
A dash marks a layer at which the effect of $\dmu$ on the second split is not positive, $S_{(2)}(\dmu)\le0$, so that the ratio is not defined: LLaMA at layer 4 for truthfulness ($S_{(2)}(\dmu)=-0.034$) and Gemma at layer 6 for sentiment ($S_{(2)}(\dmu)=-0.0005$; over both splits, $\dmu$ gives $+0.002$).
At these two layers the inequality of H1b holds trivially, because its right side is not positive; we do not count them as cases where H1b holds, since the comparison is with a direction that does not steer.
(At LLaMA layer 4 the interval of the difference lies above zero, at Gemma layer 6 it includes zero.)
Table~\ref{tab:sweep-plateau} gives the probing and steering plateaus behind the H1a verdicts, and Table~\ref{tab:sweep-h1b} the two effects behind each truthfulness ratio.

\begin{table}[h]
\centering
\small
\setlength{\tabcolsep}{2.5pt}
\resizebox{\textwidth}{!}{%
\begin{tabular}{llcccccccccccc}
\hline
 & & \multicolumn{6}{c}{Truthfulness} & \multicolumn{6}{c}{Sentiment}\\
 & & \multicolumn{2}{c}{LLaMA} & \multicolumn{2}{c}{Qwen} & \multicolumn{2}{c}{Gemma} & \multicolumn{2}{c}{LLaMA} & \multicolumn{2}{c}{Qwen} & \multicolumn{2}{c}{Gemma}\\
Depth & Layer & $P_R$ & $P_S$ & $P_R$ & $P_S$ & $P_R$ & $P_S$ & $P_R$ & $P_S$ & $P_R$ & $P_S$ & $P_R$ & $P_S$\\
\hline
0.125 & 4 / 6 & $\tau$ & $\tau$ & $\tau$ & $10\tau$--$100\tau$ & $0.1\tau$--$\tau$ & $0.1\tau$--$\dmu$ & $\tau$--$10\tau$ & $100\tau$ & $\tau$--$10\tau$ & $100\tau$ & $0.1\tau$ & $\tau$--$100\tau$\\
0.25 & 8 / 12 & $\tau$ & $100\tau$--$\dmu$ & $\tau$ & $10\tau$--$100\tau$ & $0.1\tau$--$\tau$ & $0.1\tau$--$10\tau$ & $\tau$--$10\tau$ & $100\tau$ & $\tau$--$10\tau$ & $100\tau$ & $\tau$ & $\tau$--$10\tau$\\
0.3125 & 10 / 15 & $\tau$ & $100\tau$ & $\tau$ & $100\tau$ & $0.1\tau$--$\tau$ & $100\tau$ & $10\tau$ & $100\tau$ & $10\tau$ & $100\tau$ & $\tau$ & $10\tau$\\
0.375 & 12 / 18 & $\tau$ & $100\tau$ & $\tau$ & $100\tau$ & $0.1\tau$--$\tau$ & $\dmu$ & $\tau$--$10\tau$ & $10\tau$--$100\tau$ & $10\tau$ & $100\tau$ & $\tau$ & $10\tau$\\
0.4375$^{\dagger}$ & 14 / 21 & $\tau$ & $100\tau$--$\dmu$ & $\tau$ & $100\tau$ & $\tau$ & $10\tau$--$100\tau$ & $\tau$--$10\tau$ & $100\tau$ & $\tau$--$10\tau$ & $100\tau$ & $\tau$ & $10\tau$\\
0.5 & 16 / 24 & $\tau$ & $100\tau$ & $\tau$ & $100\tau$ & $\tau$ & $100\tau$ & $10\tau$ & $100\tau$ & $\tau$--$10\tau$ & $100\tau$ & $\tau$ & $10\tau$--$100\tau$\\
0.5625 & 18 / 27 & $\tau$ & $10\tau$--$100\tau$ & $\tau$ & $100\tau$ & $\tau$ & $100\tau$--$\dmu$ & $10\tau$ & $100\tau$ & $10\tau$ & $100\tau$ & $\tau$--$10\tau$ & $100\tau$\\
0.625 & 20 / 30 & $\tau$ & $10\tau$ & $\tau$ & $10\tau$ & $\tau$ & $\dmu$ & $10\tau$ & $100\tau$ & $10\tau$ & $100\tau$ & $\tau$--$10\tau$ & $100\tau$\\
0.75 & 24 / 36 & $\tau$ & $10\tau$ & $\tau$ & $10\tau$--$100\tau$ & $\tau$ & $\dmu$ & $10\tau$ & $100\tau$ & $10\tau$ & $100\tau$ & $\tau$--$10\tau$ & $100\tau$\\
0.875 & 28 / 42 & $\tau$ & $10\tau$ & $\tau$ & $10\tau$--$100\tau$ & $\tau$ & $\dmu$ & $10\tau$ & $100\tau$ & $\tau$--$10\tau$ & $100\tau$ & $10\tau$ & $100\tau$\\
\hline
\end{tabular}}
\caption{Probing plateau $P_R$ and steering plateau $P_S$ at each layer of the sweep, on the grid $\{0.001\tau,0.1\tau,\tau,10\tau,100\tau,\dmu\}$ ($^{\dagger}$: the paper's depth). Every plateau is a contiguous range of the grid. The two plateaus overlap in five cases (truthfulness: LLaMA layer 4, Gemma layers 6 and 12; sentiment: LLaMA layer 12, Gemma layer 12); in four of them the steering plateau still ends further toward $\dmu$ than the probing plateau, so H1a holds, and at LLaMA layer 4 both plateaus are $\{\tau\}$, so it fails.}
\label{tab:sweep-plateau}
\end{table}

\begin{table}[h]
\centering
\small
\setlength{\tabcolsep}{3pt}
\begin{tabular}{llccccccccc}
\hline
 & & \multicolumn{3}{c}{LLaMA} & \multicolumn{3}{c}{Qwen} & \multicolumn{3}{c}{Gemma}\\
Depth & Layer & $\lambda^*$ & $S_{(2)}(\lambda^*)$ & $S_{(2)}(\dmu)$ & $\lambda^*$ & $S_{(2)}(\lambda^*)$ & $S_{(2)}(\dmu)$ & $\lambda^*$ & $S_{(2)}(\lambda^*)$ & $S_{(2)}(\dmu)$\\
\hline
0.125 & 4 / 6 & $\tau$ & 0.044 & -0.034 & $10\tau$ & 0.119 & 0.067 & $100\tau$ & 0.010 & 0.022\\
0.25 & 8 / 12 & $100\tau$ & 0.276 & 0.284 & $100\tau$ & 0.153 & 0.136 & $10\tau$ & 0.133 & 0.104\\
0.3125 & 10 / 15 & $100\tau$ & 0.413 & 0.411 & $100\tau$ & 0.286 & 0.242 & $100\tau$ & 0.293 & 0.227\\
0.375 & 12 / 18 & $100\tau$ & 0.398 & 0.388 & $100\tau$ & 0.372 & 0.293 & $10\tau$ & 0.277 & 0.325\\
0.4375$^{\dagger}$ & 14 / 21 & $100\tau$ & 0.310 & 0.312 & $100\tau$ & 0.647 & 0.590 & $10\tau$ & 0.587 & 0.594\\
0.5 & 16 / 24 & $100\tau$ & 0.241 & 0.189 & $100\tau$ & 0.595 & 0.569 & $100\tau$ & 0.372 & 0.334\\
0.5625 & 18 / 27 & $10\tau$ & 0.168 & 0.107 & $100\tau$ & 0.426 & 0.298 & $100\tau$ & 0.017 & 0.039\\
0.625 & 20 / 30 & $10\tau$ & 0.114 & 0.066 & $10\tau$ & 0.302 & 0.125 & $100\tau$ & 0.013 & 0.030\\
0.75 & 24 / 36 & $10\tau$ & 0.079 & 0.052 & $10\tau$ & 0.216 & 0.093 & $10\tau$ & -0.022 & 0.027\\
0.875 & 28 / 42 & $10\tau$ & 0.053 & 0.028 & $100\tau$ & 0.189 & 0.143 & $10\tau$ & -0.028 & 0.030\\
\hline
\end{tabular}
\caption{The two effects behind each truthfulness ratio of Table~\ref{tab:sweep-steer}: the direction $\lambda^*$ chosen on the first held-out split, and the effects of $\lambda^*$ and of $\dmu$ on the second split (change of the answer margin in nats at $K=\kappa_0$; $^{\dagger}$: the paper's depth). The sentiment ratios divide effects between 0.04 and 1.95 and lie between 1.23 and 1.97 at every defined layer, so we do not list them.}
\label{tab:sweep-h1b}
\end{table}

H1a holds in 59 of the 60 combinations of model, concept and layer (Table~\ref{tab:sweep-plateau}).
The exception is LLaMA at layer 4 for truthfulness, where $\dmu$ does not steer at all and both plateaus are $\{\tau\}$.
For truthfulness the probing plateau is $\tau$ at every layer of LLaMA and Qwen and $\tau$ or $0.1\tau$--$\tau$ in Gemma, and the steering plateau lies between $10\tau$ and $\dmu$ at every layer except three low ones (LLaMA layer 4, Gemma layers 6 and 12).
For sentiment the probing plateau lies at $\tau$--$10\tau$ and the steering plateau at $100\tau$ (in Gemma, $10\tau$--$100\tau$).
For sentiment, H1b holds at every layer at which $\dmu$ steers.
For truthfulness, H1b depends on the layer: it fails at the layers we use, as in Section~\ref{sec:res-rq1}, but holds at layers 16 and 18 of LLaMA, at layers 10, 12 and 18 to 28 of Qwen, and at layer 15 of Gemma (Table~\ref{tab:sweep-steer}).
The ratios above 2 in Qwen (layers 20 and 24) and the negative ratios in Gemma (layers 36 and 42) both divide by a small effect of $\dmu$ (Table~\ref{tab:sweep-h1b}).
In Qwen, $\lambda^*=10\tau$ moves the answer margin by 0.302 and 0.216 nats against 0.125 and 0.093 for $\dmu$, and the intervals of the difference lie above zero.
In Gemma, $\lambda^*=10\tau$, chosen on the first split, moves the margin the wrong way on the second split ($-0.022$ and $-0.028$) while $\dmu$ moves it by $0.027$ and $0.030$, and the intervals lie below zero; at these layers $\dmu$ steers truthfulness at less than a tenth of its effect at layer 21, so the ratio compares two small numbers and the sign of $\lambda^*$ is not stable across splits.

The effect of $\dmu$ on truthfulness is largest at mid depth, at layer 10 in LLaMA, 14 in Qwen and 21 in Gemma, and falls toward zero in the upper layers (0.011 at layer 28 of LLaMA and below 0.05 from layer 27 of Gemma), where ratios to $\dmu$ divide by small numbers.
The effect on sentiment rises to about half depth and then stays level in LLaMA, keeps rising in Qwen, and is largest at layer 24 in Gemma.
Because $\kappa_0$ is set separately at each layer (for truthfulness in LLaMA, from 0.0004 at layer 4 to 0.0062 at layer 28), effects at different layers are read at different side effects; within a layer, all directions share the same $\kappa_0$.

The concept difference itself holds at the depths where $\dmu$ steers truthfulness: at layers 8 to 20 of LLaMA and 8 to 18 of Qwen, the best probe of the sweep steers truthfulness at only 16--70\% of the effect of $\dmu$, while for sentiment it steers at 71--172\% of it at every layer of every model.
At the upper layers, and at most layers of Gemma, $\dmu$ steers truthfulness too weakly for this ratio to be informative.

We therefore keep the layers chosen in advance.
Probing varies little from $3/8$ of the depth upward, and H1a holds at every layer except the lowest one of LLaMA for truthfulness. The truthfulness verdict of H1b holds at mid depth, where $\dmu$ steers truthfulness most strongly, and not at every layer. The sweep covers RQ1 only.
The PopQA column of Table~\ref{tab:sweep-probe} is a probing sweep; the steering of RQ2 (the gated $\dmu$ of Section~\ref{sec:res-rq2}, including the H2b(iii) and H2c verdicts) was run at the paper's layer only, so those verdicts are single-layer results and we are not aware about how they change with depth.

\section{Additional Results and Figures}
\label{app:figs}

This appendix provides additional figures supporting the results in Section~\ref{sec:results}.
Each figure is followed by a short description of what it shows.

\subsection{Probing and steering within models}

\begin{figure}[!htbp]
\centering
\includegraphics[width=\linewidth]{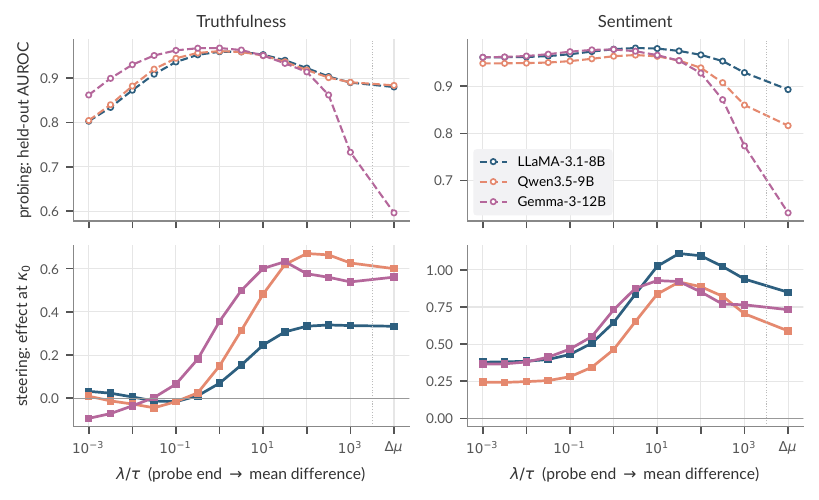}
\caption{Probing (top, held-out AUROC) and steering at $\kappa_0$ in each concept's own units (bottom) against $\lambda/\tau$. The shapes match Figure~\ref{fig:rq1-dial}, but the size of the effects differs between models.}
\label{fig:app-dial-raw}
\end{figure}

Gemma shows the split most clearly: its mean difference hardly separates the classes (AUROC 0.596 for truthfulness and 0.631 for sentiment), yet steering along it still changes the model's behavior: a direction can be nearly useless as a probe and still steer.

Figure~\ref{fig:rq1-dial} divides every steering effect by that of $\dmu$, which makes the models comparable but hides how large the effects are.
Figure~\ref{fig:app-dial-raw} shows the raw values against $\lambda/\tau$, with the steering effect in nats of answer margin for truthfulness and as the change of the log-odds of positive versus negative words for sentiment.
The shape is the same in all three models: probing peaks near $\lambda=\tau$, while steering rises from almost nothing at the probe end to its maximum further to the right.
For truthfulness, the probe end barely steers at all; the effect stays close to zero up to about $\lambda=0.1\tau$.
The models differ in how much they move: for truthfulness, Qwen and Gemma move the answer margin about twice as much as LLaMA at the same side effect, and for sentiment, LLaMA moves the most.
The steep drop of Gemma's AUROC toward $\dmu$ is the near-chance probing of its mean difference discussed above.

\begin{figure}[!htbp]
\centering
\includegraphics[width=\linewidth]{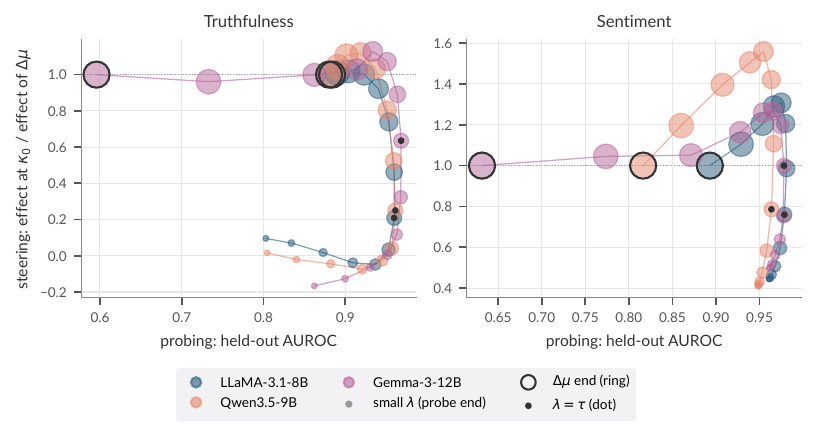}
\caption{Each direction of the family at its probing AUROC (horizontal) and its steering effect relative to $\dmu$ (vertical). No curve ends in the upper right corner, so no direction is best at both.}
\label{fig:app-plane}
\end{figure}

In Figure~\ref{fig:app-plane}, lines connect the directions in the order of the family, bubbles grow from the probe end to $\dmu$, the ring marks $\dmu$, the small dot marks $\lambda=\tau$, and the dashed line marks the steering effect of $\dmu$.
If one direction were best at both tasks, the curve of each model would end in the upper right corner.
It does not.
For truthfulness, each curve runs up along the right edge and then bends to the left: the directions that probe best (AUROC about 0.96) steer weakly, and the directions that steer best probe somewhat worse.
For sentiment, the strongest steering (up to 1.6 times that of $\dmu$, for Qwen) comes from directions that probe well but not best, and $\dmu$ itself lies to the left of them.
Gemma's ring sits near an AUROC of 0.6 in both panels: its mean difference steers, but hardly probes.

\subsection{Probing and steering across models}
\label{app:figs-xmodel}

Figure~\ref{fig:xmodel} shows the H1c comparison of Section~\ref{sec:res-rq1}.
As controls, we also steer with $\dmu_A$ carried by the probe law, $M_{B\rightarrow A}^{\top}\dmu_A$, probe with the transferred direction under the steering law and under an orthogonal Procrustes map (used for probing only; all other transfer results use the ridge maps), steer with the source's probe end carried by the steering law, and compare $\cos(M_{A\rightarrow B}\dmu_A,\dmu_B)$ with a null in which $A$'s labels are permuted; Table~\ref{tab:xmodel-controls} gives these per pair.

Each point is one ordered pair source$\rightarrow$target (L = LLaMA, Q = Qwen, G = Gemma) for one concept, and the dashed diagonal marks equal ratios.
The shaded corners are probe-shared ($\rho_{\mathrm{probe}}\ge0.8$, $\rho_{\mathrm{steer}}<0.5$), steering-shared (the reverse), both-shared (both $\ge0.8$) and neither-shared (both $<0.5$).
All pairs lie at the right edge, but most lie below the diagonal.

Three caveats temper this reading.
First, $\rho_{\mathrm{probe}}$ and $\rho_{\mathrm{steer}}$ are ratios of different quantities, an AUROC above chance and a steering effect, so comparing them is only approximate.
In particular, AUROC saturates: under the Gaussian model of Appendix~\ref{app:witness}, $\mathrm{AUROC}=\Phi(d'/\sqrt2)$ with $d'$ the standardized separation, and at a target AUROC of 0.96--0.98 a probe that loses a third of its $d'$ still keeps about 85\% of the AUROC above chance.
Recomputed on the $d'$ scale from the same held-out AUROCs, $\rho_{\mathrm{probe}}$ falls to 0.65--0.79 and exceeds $\rho_{\mathrm{steer}}$ in 9 of 12 pairs instead of 11: all six truthfulness pairs hold, while for sentiment LLaMA$\rightarrow$Qwen (0.76 against 0.93), LLaMA$\rightarrow$Gemma (0.66 against 0.72) and Gemma$\rightarrow$Qwen (0.79 against 0.81) no longer do.
For ``unknown'', whose target probes lie further from the ceiling (AUROC 0.92--0.94), the $d'$ ratio is 0.82--0.92 against a steering ratio of 0.30--0.79, so all six pairs hold on either scale.

Second, the maps are fitted on neutral text and fit the concept texts less well ($R^2$ of 0.13--0.47, against 0.48--0.70 on held-out neutral text; Table~\ref{tab:xmodel-controls}).
Part of the loss in steering may therefore come from the map rather than from the models, and for the same reason no pair can be called ``neither shared''.

Third, $\rho_{\mathrm{probe}}$ and $\rho_{\mathrm{steer}}$ measure what a carried direction does in the target, how well it reads the concept and how strongly it pushes the model, not how close it lies to the target's own direction; a carried direction can do the job while pointing away from the target's own $\dmu$.
The measure is therefore functional rather than geometric.
In 10 of 12 pairs, the mapped $\dmu$ is closer to the target's own $\dmu$ than a mean difference computed with permuted labels (cosine 0.53--0.81, against a 95th null percentile of 0.12--0.66).
In the two sentiment pairs into Gemma, however, it is no closer to Gemma's own $\dmu$ than the permuted-label null (cosine $-0.17$ and $-0.02$) and still steers at 0.72 and 0.70 of its effect.

The disagreement is confined to the mean-difference end.
Along the family, the carried source direction $M_{A\rightarrow G}w_{A,\lambda}$ and Gemma's own $w_{G,\lambda}$ have a positive cosine that rises from 0.20 at the probe end to 0.50 at $32\tau$ (from LLaMA) and 0.64 at $100\tau$ (from Qwen), and turns negative only at $\dmu$.

Along the family, the carried source direction $M_{A\rightarrow G}w_{A,\lambda}$ and Gemma's own $w_{G,\lambda}$ have a positive cosine that rises from 0.20 at the probe end to 0.50 at $32\tau$ (from LLaMA) and 0.64 at $100\tau$ (from Qwen), and turns negative only at $\dmu$.
Gemma's own sentiment $\dmu$ probes at an AUROC of 0.63 (Appendix~\ref{app:figs}), so most of its length lies along directions that do not carry the concept; the carried $\dmu_A$ points away from those directions and with the regularized ones that carry it, which is probably why it steers the right way.
The same holds for ``unknown'' on PopQA (Appendix~\ref{app:rq2}).

\begin{table}[h]
\centering
\small
\setlength{\tabcolsep}{3.5pt}
\resizebox{\textwidth}{!}{%
\begin{tabular}{llcccccccccc}
\hline
 & & \multicolumn{2}{c}{Map $R^2$} & \multicolumn{3}{c}{$\rho_{\mathrm{probe}}$ at $B$'s best $\lambda$} & \multicolumn{2}{c}{$\rho_{\mathrm{steer}}$ of $\dmu_A$} & \multicolumn{2}{c}{$\rho_{\mathrm{steer}}$, steering law}\\
Concept & Pair & neutral & concept & probe law & steering law & Procrustes & steering law & probe law & $w_{A,\tau}$ & $w_{A,10\tau}$\\
\hline
Truthfulness & L$\to$Q & 0.56 & 0.16 & 0.81 & 0.62 & 0.80 & 0.64 & 0.61 & 0.15 & 0.41\\
 & L$\to$G & 0.70 & 0.26 & 0.89 & 0.10 & 0.89 & 0.44 & 0.30 & 0.20 & 0.42\\
 & Q$\to$L & 0.50 & 0.14 & 0.83 & 0.74 & 0.83 & 0.54 & 0.39 & 0.40 & 0.71\\
 & Q$\to$G & 0.69 & 0.25 & 0.88 & 0.10 & 0.90 & 0.53 & 0.21 & 0.27 & 0.49\\
 & G$\to$L & 0.48 & 0.15 & 0.91 & 0.83 & 0.93 & 0.62 & -0.13 & 0.56 & 0.74\\
 & G$\to$Q & 0.53 & 0.13 & 0.89 & 0.87 & 0.90 & 0.69 & -0.03 & 0.39 & 0.67\\
\hline
Sentiment & L$\to$Q & 0.56 & 0.33 & 0.89 & 0.84 & 0.95 & 0.93 & 0.85 & 0.68 & 1.10\\
 & L$\to$G & 0.70 & 0.47 & 0.86 & -0.10 & 0.57 & 0.72 & 0.32 & 0.60 & 0.80\\
 & Q$\to$L & 0.50 & 0.24 & 0.91 & 0.86 & 0.95 & 0.65 & 0.45 & 0.74 & 0.88\\
 & Q$\to$G & 0.69 & 0.37 & 0.91 & 0.17 & 0.76 & 0.70 & 0.28 & 0.63 & 0.79\\
 & G$\to$L & 0.48 & 0.23 & 0.91 & 0.86 & 0.93 & 0.70 & 0.05 & 0.74 & 0.87\\
 & G$\to$Q & 0.53 & 0.28 & 0.91 & 0.85 & 0.93 & 0.81 & 0.07 & 0.95 & 1.20\\
\hline
\end{tabular}
}
\caption{Cross-model controls per ordered pair $A\to B$ (L = LLaMA, Q = Qwen, G = Gemma). Map $R^2$: held-out $R^2$ of $M_{A\rightarrow B}$ on neutral text and on the concept texts. $\rho_{\mathrm{probe}}$: AUROC above chance of the transferred direction at $B$'s best $\lambda$, relative to $B$'s own, under the probe law $M_{B\rightarrow A}^{\top}w_A$ (the value of Figure~\ref{fig:xmodel}), the steering law $M_{A\rightarrow B}w_A$, and the orthogonal Procrustes map $Qw_A$. $\rho_{\mathrm{steer}}$: steering effect in $B$ at its $\kappa_0$ relative to $B$'s own $\dmu$, for $\dmu_A$ under the steering law (the value of Figure~\ref{fig:xmodel}) and the probe law, and for the source's directions $w_{A,\tau}$ and $w_{A,10\tau}$ under the steering law. Steering with the Procrustes map was not run.}
\label{tab:xmodel-controls}
\end{table}

Table~\ref{tab:xmodel-controls} gives the controls per pair.
The orthogonal Procrustes map, fitted on the same neutral text, probes about as well as the probe law in 10 of the 12 pairs (0.80--0.95 against 0.81--0.91) and worse only in the two sentiment pairs into Gemma (0.57 and 0.76 against 0.86 and 0.91), whereas probing under the steering law breaks down in every pair into Gemma (0.10, 0.10, $-0.10$ and 0.17).
The loss under the steering law therefore comes from applying $M_{A\rightarrow B}$ forward to a probe, not from the maps being fitted on neutral text.
We did not steer with the Procrustes map, so it enters only the probing comparison.
The source's probe end $w_{A,\tau}$, carried by the steering law, steers at 0.15--0.56 of the target's $\dmu$ for truthfulness and 0.60--0.95 for sentiment, below the transferred $\dmu_A$ in 10 of 12 pairs; $w_{A,10\tau}$ steers at 0.41--0.74 and 0.79--1.20.
$w_{A,\tau}$ lies on the source's probing plateau for truthfulness in all three models and for sentiment in Gemma, and one grid step below it ($3\tau$) for sentiment in LLaMA and Qwen; we did not steer with transferred probe directions under the probe law.
The concept $R^2$ is lowest for truthfulness (0.13--0.26) and higher for sentiment (0.23--0.47), while the neutral $R^2$ depends only on the pair of models (highest into Gemma, 0.69--0.70; lowest into LLaMA, 0.48--0.50).

\begin{figure}[!htbp]
\centering
\includegraphics[width=0.72\linewidth]{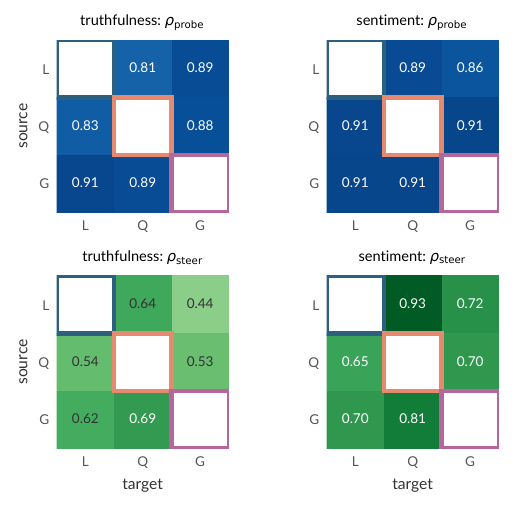}
\caption{$\rho_{\mathrm{probe}}$ (top) and $\rho_{\mathrm{steer}}$ (bottom) for each source (rows) and target (columns); darker cells are higher. Probing values are uniformly high, steering values lower and uneven.}
\label{fig:app-matrices}
\end{figure}

Figure~\ref{fig:app-matrices} gives the values of Figure~\ref{fig:xmodel} as tables, with an empty diagonal because a model is not carried into itself.
The probing values are uniformly high (0.81--0.91) and look much the same for every pair of models.
The steering values are lower and uneven.
Steering directions carried into Qwen work best (0.64 and 0.69 for truthfulness, 0.93 and 0.81 for sentiment), and those carried into Gemma work worst for truthfulness (0.44 and 0.53).

\begin{figure}[!htbp]
\centering
\includegraphics[width=\linewidth]{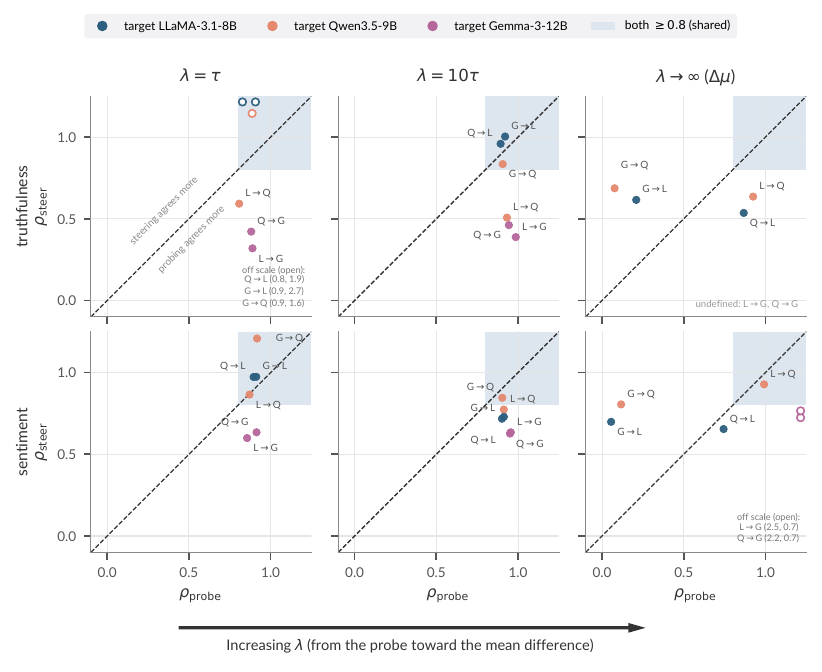}
\caption{$\rho_{\mathrm{probe}}$ (horizontal) and $\rho_{\mathrm{steer}}$ (vertical), both computed at $\lambda=\tau$ (left), $10\tau$ (middle) and $\dmu$ (right). Probes transfer well except at the $\dmu$ end with Gemma, and steering transfer depends on the point of the family.}
\label{fig:app-grid}
\end{figure}

Figure~\ref{fig:xmodel} uses the comparison fixed in advance: the probe at the target's best $\lambda$, and steering with $\dmu$.
Figure~\ref{fig:app-grid} repeats the comparison at fixed points of the family, with both ratios computed at the same $\lambda$.
Each point is one source$\rightarrow$target pair, coloured by the target; the dashed diagonal marks equal ratios, the shaded corner marks ``both shared'', and values outside the plotted range are listed in the panel.
Probes transfer well near the probe end and at $10\tau$, with $\rho_{\mathrm{probe}}$ between about 0.8 and 1.0.
At the $\dmu$ end, probe transfer breaks down whenever Gemma is involved. Gemma's own $\dmu$ probes poorly (AUROC 0.596 for truthfulness and 0.631 for sentiment), so ratios into Gemma are either not defined (truthfulness; our rule requires the target's own AUROC to be at least 0.6, and 0.596 falls just below it) or above 2 because they divide by a small number (sentiment), and Gemma's $\dmu$ carried into another model reaches an AUROC of only 0.52--0.58.
Steering transfer depends more on where on the family we look.
At $\lambda=\tau$, three truthfulness ratios (1.6--2.7 for Q$\to$L, G$\to$L and G$\to$Q) exceed the plotted range and are listed in the panel, because the target's own probe end steers weakly (0.07 nats in LLaMA and 0.15 in Qwen, against 0.33 and 0.60 for their $\dmu$) and the ratio divides by a small number.
At $10\tau$, the sentiment pairs lie between 0.63 and 0.85, lowest for the two pairs into Gemma, while the truthfulness pairs split into two groups: close to 1 for the pairs into LLaMA and from Gemma into Qwen, and 0.4--0.5 for the other three.

\begin{figure}[!htbp]
\centering
\includegraphics[width=\linewidth]{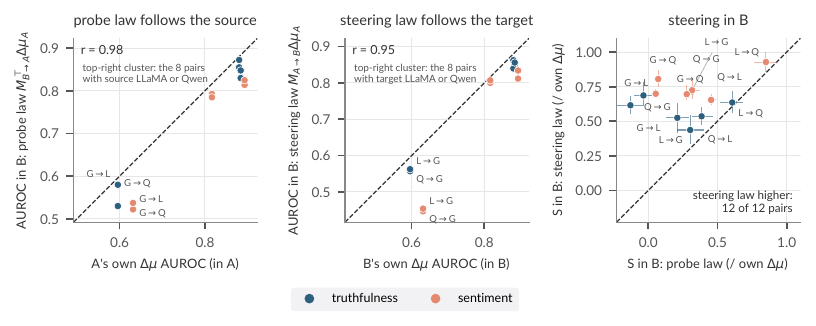}
\caption{Probing after the probe law (left) and after the steering law (middle), and steering after each law (right), for each model pair and concept, with the pairs that involve Gemma labelled in the first two panels. The probe law keeps what a direction reads, and the steering law keeps where it pushes.}
\label{fig:app-laws}
\end{figure}

The two laws carry different things (Figure~\ref{fig:app-laws}).
The left panel plots the probing AUROC in the target after the probe law against the source's own $\dmu$ AUROC, and the middle panel plots the AUROC after the steering law against the target's own $\dmu$ AUROC; the right panel plots the steering effect at the target's $\kappa_0$, relative to its own $\dmu$, after the probe law (horizontal) and the steering law (vertical), with 95\% intervals. Dashed lines mark equality, and $r$ is the Pearson correlation over the twelve pairs.
After the probe law, how well $\dmu_A$ probes in the target follows how well it probed in the source ($r=0.98$ across the twelve pairs): the probe law keeps what the direction reads.
After the steering law, it follows how well the target's own $\dmu$ probes ($r=0.95$): the steering law keeps where the direction pushes, and in the new model that push looks like the target's own mean difference.
For steering, the steering law wins in all twelve pairs.
Gemma makes the difference visible.
Its $\dmu$ probes poorly in Gemma and, using the probe law, also in LLaMA and Qwen, where it does not steer at all; using the steering law, it steers 0.62--0.81 as strongly as their own $\dmu$.
Both correlations are driven largely by the contrast between Gemma and the other two models, whose AUROCs are similar.
A probe and a steering direction are different objects, and they have to be mapped differently.

\subsection{Answer reachability and abstention}

\begin{figure}[!htbp]
\centering
\includegraphics[width=\linewidth]{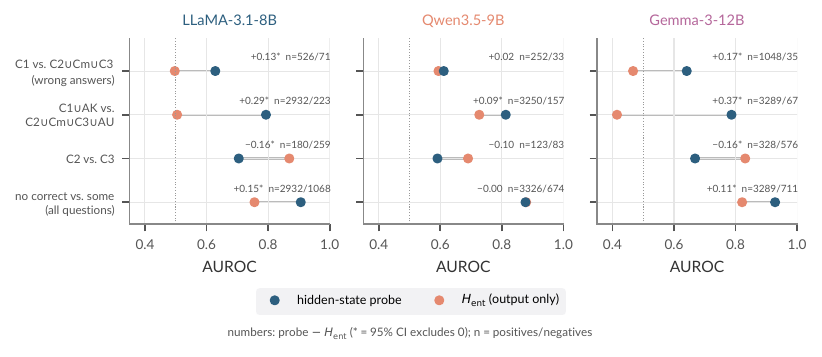}
\caption{AUROC of the hidden-state probe (dark) and of the first-token entropy $H_{\mathrm{ent}}$ (light) for the three contrasts of H2a and, in the last row, for all questions; the dotted line marks chance. The probe beats the entropy once abstentions are included, and the entropy beats the probe on C2 against C3.}
\label{fig:app-rq2-read}
\end{figure}

The numbers in Figure~\ref{fig:app-rq2-read} give the probe minus $H_{\mathrm{ent}}$ (* if the 95\% interval excludes zero) and the number of questions on each side.
Among wrong answers only (first row), the probe beats the entropy in LLaMA and Gemma (+0.13 and +0.17) but not in Qwen, and all AUROCs are low (0.61--0.64), partly because C1 is small.
Once abstentions are included (second row), the probe is clearly better in all three models (+0.09 to +0.37), while the entropy is close to or below chance in LLaMA and Gemma.
Uncertain and confident errors (third row) are separated better by the entropy (0.69--0.87) than by the probe (0.59--0.71), significantly in LLaMA and Gemma.

The second route of the rule, a cross-fitted logistic combination of the probe and $H_{\mathrm{ent}}$, changes no verdict: against $H_{\mathrm{ent}}$ alone, the combination gains $+0.12$ $[+0.01,+0.20]$, $-0.07$ $[-0.17,+0.04]$ and $+0.17$ $[+0.05,+0.27]$ AUROC on C1 against C2\,$\cup$\,Cm\,$\cup$\,C3 in LLaMA, Qwen and Gemma, $+0.30$, $+0.16$ and $+0.38$ with abstentions included (all intervals above zero), and $-0.01$ $[-0.02,-0.00]$, $+0.02$ $[-0.05,+0.09]$ and $-0.03$ $[-0.04,-0.02]$ on C2 against C3.
Over all questions the combination exceeds the entropy in Qwen ($+0.07$ $[+0.06,+0.08]$), where the probe alone equals it.

The last row covers all questions and compares those with no correct answer among the six with those with at least one, the labels the probe is trained on. There, the probe is better than the entropy in LLaMA and Gemma (+0.15 and +0.11) and equal in Qwen (0.877 against 0.879).

%\begin{figure}[!htbp]
%\centering
%\includegraphics[width=\linewidth]{figures/fig_rq2_plane.pdf}
%\caption{The trade-off of each steering direction. Goal: compare the directions by %how much abstention they gain on errors and how much they cost on correct answers.
%Horizontal: abstention gained on errors without a correct answer; vertical: %abstention on questions the model answered correctly, which is the cost; both in %percentage points against the unsteered rerun.
%Each marker is one direction at one strength (larger markers are stronger); black %outlines mark the strength at which $\dmu$ reaches $\kappa_0$.
%Lower right is better.}
%\label{fig:app-rq2-plane}
%\end{figure}

%Within each model, the directions in Figure~\ref{fig:app-rq2-plane} fall roughly on one line: every extra point of abstention on questions without a correct answer costs a similar amount on correct answers, whichever direction is used.
%The random direction gains little.
%The probe-gated $\dmu$ lies below the line of $\dmu$ in all three models: at the same gain, it lowers the cost by 34 to 59\% (LLaMA: 8.6 against about 13.1 points at a gain of about 38 points; Qwen: 4.7 against about 7.4 at a gain of about 33; Gemma: 2.5 against about 6.1 at a gain of about 32).
%This comparison at the same gain was not planned in advance, unlike H2c, which compares at the same strength and fails in LLaMA and Gemma.

\begin{figure}[!htbp]
\centering
\includegraphics[width=0.9\linewidth]{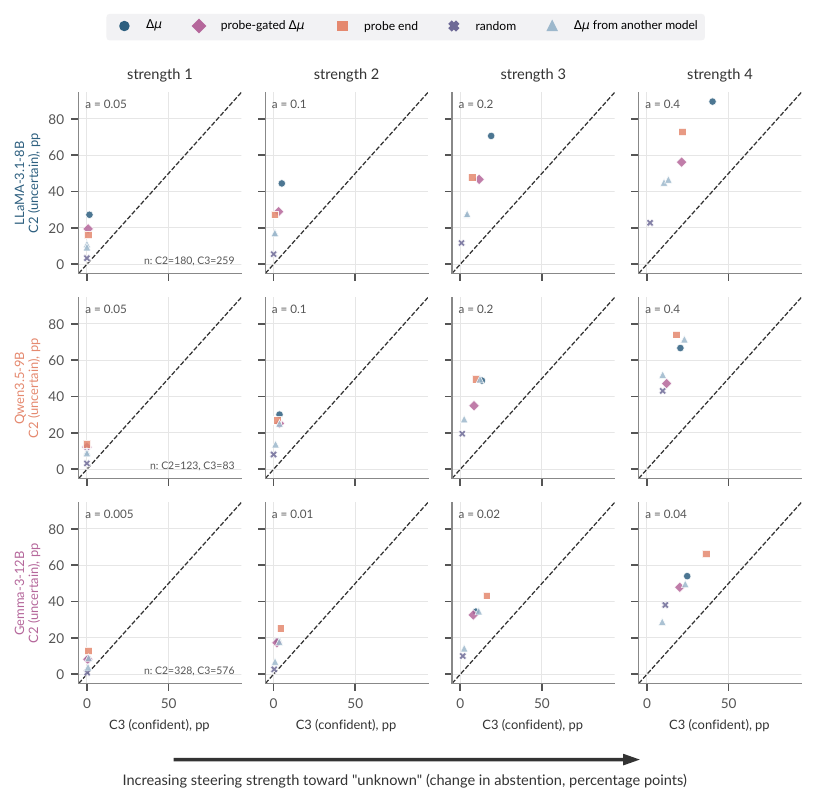}
\caption{Change in abstention (pp) on confident errors (C3, horizontal) and uncertain errors (C2, vertical) for each model (rows) and strength (columns). All points lie above the diagonal.}
\label{fig:app-rq2-grid}
\end{figure}

At every strength and for every direction, including the random one, the points in Figure~\ref{fig:app-rq2-grid} lie above the diagonal: uncertain errors turn into abstentions more easily than confident ones.
Confident errors resist: they start to move only at strengths where correct answers also start to abstain (for LLaMA at $a=0.4$, +40 on C3 and +33 on C0).

\end{document}